\documentclass{article}

\usepackage[final]{neurips_2024}

\usepackage[utf8]{inputenc} 
\usepackage[T1]{fontenc}    
\usepackage{hyperref}       
\usepackage{url}            
\usepackage{booktabs}       
\usepackage{amsfonts}       
\usepackage{nicefrac}       
\usepackage{microtype}      
\usepackage{xcolor}         

\usepackage{amsmath,amssymb}
\usepackage{commath}
\usepackage{bm}
\usepackage{algorithm}
\usepackage{algorithmic}
\usepackage{multirow}
\usepackage{graphicx} 
\usepackage{subcaption}
\usepackage[capitalize]{cleveref}
\setcitestyle{numbers,square}
\hypersetup{
    colorlinks=true,    
    linkcolor=blue,     
    citecolor=blue,     
    filecolor=blue,     
    urlcolor=blue       
}

\renewcommand{\algorithmicrequire}{\textbf{Input:}}  
\renewcommand{\algorithmicensure}{\textbf{Output:}} 

\title{Conjugate Bayesian Two-step Change Point Detection for Hawkes Process}

\author{Zeyue Zhang$^{1,2}$, Xiaoling Lu$^{1,2}$, Feng Zhou$^{1,3}$\thanks{Corresponding author.}\\
$^1$Center for Applied Statistics and School of Statistics, Renmin University of China\\
$^2$Innovation Platform, Renmin University of China\\
$^3$Beijing Advanced Innovation Center for Future Blockchain and Privacy Computing\\
\texttt{ \{zhangzeyue, xiaolinglu, feng.zhou\}@ruc.edu.cn}
}

\begin{document}

\maketitle

\begin{abstract}
  The Bayesian two-step change point detection method is popular for the Hawkes process due to its simplicity and intuitiveness. However, the non-conjugacy between the point process likelihood and the prior requires most existing Bayesian two-step change point detection methods to rely on non-conjugate inference methods. These methods lack analytical expressions, leading to low computational efficiency and impeding timely change point detection.
To address this issue, this work employs data augmentation to propose a conjugate Bayesian two-step change point detection method for the Hawkes process, which proves to be more accurate and efficient. Extensive experiments on both synthetic and real data demonstrate the superior effectiveness and efficiency of our method compared to baseline methods. Additionally, we conduct ablation studies to explore the robustness of our method concerning various hyperparameters. Our code is publicly available at \url{https://github.com/Aurora2050/CoBay-CPD}. 
\end{abstract}

\section{Introduction}

Point process data, characterized by a series of discrete events occurring over time, finds extensive applications in various domains, including finance~\citep{bacry2015hawkes}, neuroscience~\citep{zhou2021efficient}, and social networks~\citep{pinto2015trend}.
Hawkes processes~\citep{hawkes1971spectra}, a subclass of point processes, have gained attention due to their ability to model self-exciting and clustering behaviors. However, traditional modeling of Hawkes processes relies on the assumption that the parameters of the process, i.e., the distribution of the process, remain invariant over time. In practice, this assumption often fails to hold, as the underlying dynamics of point process data can change over time~\citep{wang2023sequential,zhou2020fast}. 

The violation of the time-invariant parameter assumption poses a significant challenge in accurately modeling Hawkes process data. Real-world processes are inherently dynamic and subject to various influences, leading to fluctuations in process distribution. For instance, in a social media platform, the intensity of user interactions may change due to a sudden surge in activity during an emergency event or a significant drop in user engagement during a service outage. To address this issue, the change point detection (CPD) has emerged as a pivotal task in analyzing point process data. CPD aims to identify the locations where the parameters (distribution) of the process undergo a significant change. 

In this work, our focus is on CPD within Hawkes process. Previous studies have tackled this issue, offering diverse methodologies for this purpose~\citep{bhaduri2021change,detommaso2019stein,wang2023sequential,zhou2020fast}. Among these, the two-step estimation-prediction method~\citep{detommaso2019stein} has gained widespread application due to its simplicity and intuitiveness. This method involves using historical data to estimate model parameters and then utilizing these estimated parameters to estimate the distribution of the next event point. 
If the observed point aligns closely with the prediction, model parameters are assumed unchanged. But if there's a notable deviation, it suggests parameter change, indicating a current change point. 

An issue with the estimation-prediction method lies in the inaccurate estimation of model parameters. 
This issue stems from the constraints within the CPD system: 
on one hand, for efficient detection, we cannot use a large amount of historical data as it would lead to a heavy computational burden; 
on the other hand, once a change point is identified, the reliance on samples post-change point for estimating subsequent model parameters restricts the availability of adequate data for initial post-change point estimations. 
This often results in inaccurate parameter estimations due to our reliance on a limited historical dataset, subsequently affecting the prediction of the next event point. As a consequence, the algorithm may wrongly identify many non-change points as change points (false positives) and vice versa (false negatives). 

To alleviate the aforementioned issues, many studies propose utilizing Bayesian approaches. Compared to frequentist methods, Bayesian methods exhibit better robustness as they not only consider samples but also incorporate prior knowledge. 
When data is limited, prior knowledge acts as regularization, effectively preventing overfitting. 
Specifically, the Bayesian estimation-prediction method estimates the posterior distribution of model parameters based on historical data and then leverages this posterior distribution to estimate the predictive distribution of the next event point. This predictive distribution considers all possible model specifications, which differs from the frequentist method. 
Similarly, if the observed point closely matches the prediction, it indicates no change point, suggesting unchanged model parameters. Conversely, a significant deviation suggests a change point\footnote{Here, ``Bayesian'' refers to the Bayesian treatment of model parameters; for the change point, it remains a point estimation.}. 

Past studies have investigated the Bayesian Hawkes process~\citep{linderman2017bayesian,rasmussen2013bayesian}. Due to the non-conjugacy between point process likelihoods and any priors, inferring the posterior of Hawkes process parameters presents significant challenges. 
Currently, the majority of work utilizes methods such as Markov chain Monte Carlo (MCMC)~\citep{neal1993probabilistic} or variational inference~\citep{blei2017variational} to infer the posterior in non-conjugate scenarios. 
Inference methods derived in such non-conjugate scenarios often lack analytical expressions, leading to low computational efficiency.
Therefore, they are not well-suited for the CPD system, which requires timeliness. 

To address this challenge, this paper employs a data augmentation strategy recently proposed in the Bayesian point process field~\citep{donner2018efficient,malem2021flexible,zhou2020auxiliary,zhou2021efficient,zhou2022efficient}. This strategy augments the Hawkes process likelihood with auxiliary latent variables, enabling the augmented Hawkes process likelihood conditionally conjugate to the prior. Leveraging the conditionally conjugate model, we can derive an analytical Gibbs sampler that enables closed-form iterative sampling. 
The effectiveness and efficiency of our proposed method are demonstrated through experiments using both synthetic and real data. 

Specifically, we make the following contributions: 
\textbf{(1)} We propose the conjugate Bayesian two-step change point
detection (CoBay-CPD) for Hawkes process, which leverages data augmentation to address the non-conjugate issue. This novel method allows for more accurate and efficient CPD in Hawkes process. 
\textbf{(2)} We develop an analytical Gibbs sampler tailored for the proposed model, enabling closed-form iterative sampling of the model parameters. This streamlines the inference process and alleviates the computational burden associated with non-conjugate scenarios. 
\textbf{(3)} 
The experiments demonstrate that our method achieves accurate and timely detection of change points in Hawkes process compared to baseline models, which highlights its practical applicability across various dynamic event modeling scenarios.

\section{Related Works}
In this section, we delve into the existing literature concerning CPD and Bayesian inference for Hawkes processes. 

\subsection{Change Point Detection for Hawkes Process}
CPD in Hawkes processes has remained an exceptionally challenging endeavor. Within the frequentist framework, several methods have been proposed. For instance, techniques based on second-order statistics have been introduced \citep{zhou2020fast}, as well as approaches leveraging the cumulative sum (CUSUM) method \citep{wang2023sequential} and sequential testing strategies \citep{bhaduri2021change}. However, Bayesian approaches have remained underexplored. A pioneering contribution by~\cite{detommaso2019stein} introduced a Bayesian approach based on Stein variational inference~\citep{liu2016stein} tailored for Hawkes processes. 
However, this method derived in non-conjugate scenarios lacks analytical expressions, leading to low computational efficiency. Subsequent experiments demonstrate that our CoBay-CPD significantly enhances the computational efficiency. 

\subsection{Bayesian Inference for Hawkes Process}
The Bayesian Hawkes process has been a hot topic in research, broadly categorized into two classes: parametric and non-parametric methods. 
Parametric approaches~\citep{linderman2015scalable,rasmussen2013bayesian} involve applying priors to parameters in Hawkes process and inferring the posterior. 
Non-parametric methods~\citep{rui2019variational,zhang2018efficient,zhou2021efficientem}, offering greater flexibility than parametric ones, model the background rate and influence function (refer to \cref{hawkes_intensity}) as flexible functions, apply priors (commonly Gaussian processes) on these functions, and aim to infer the posterior of these functions. 
As the point process likelihood is not conjugate to any priors, both parametric and non-parametric approaches face challenges in posterior inference. 
Most studies rely on methods such as MCMC, variational inference, or Laplace approximation~\citep{mackay2003information}. 

\subsection{Data Augmentation for Hawkes Process}
In recent years, the Bayesian point process field has introduced a novel data augmentation technique to address non-conjugate inference challenges. This method introduces auxiliary latent variables into point process likelihood, transforming non-conjugate problems into conditionally conjugate ones, and thus enabling the derivation of fully analytical inference algorithms. Parametric studies are referenced in~\cite{zhou2021efficient,zhou2022efficient}, while non-parametric research is detailed in~\cite{malem2022variational,sulem2022scalable,zhou2020auxiliary}. 
With analytical expressions, the inference algorithms based on data augmentation exhibit higher computational efficiency than the non-analytical ones derived in non-conjugate scenarios. Our study adopts the data augmentation, opting for a computationally more efficient parametric approach to ensure the efficiency of CPD.

\section{Methodology}
In this section, we present our proposed CoBay-CPD method for Hawkes process. 
We outline its key derivation steps while providing further details in \cref{app_gibbs}. 

\subsection{Hawkes Process with Inhibition}
\label{hawkes_inhibition}
The mathematical foundation of Hawkes process is defined by the conditional intensity function that represents the instantaneous rate of event occurrences at time $t$ given the history up to but not including $t$. It is defined as follows: 
\begin{equation}
\lambda^*(t) =\lambda(t|\mathcal{H}_{t^-})=\mu+\sum_{t_i<t}\phi(t-t_i), \label{hawkes_intensity}
\end{equation}
where $\mu$ is the background rate, $\phi(\cdot)$ is the influence function representing the self-excitation effect from event occurring at $t_i$ to $t$, the summation captures the influence of all past events, $\mathcal{H}_{t^-}$ is the historical information up to but not including $t$, and $*$ indicates the intensity depends on the history. 
The self-exciting property of Hawkes processes allows events to trigger additional events, leading to clustering and bursty behavior in event sequences.

Traditional Hawkes processes only employ positive influence functions to avoid negative intensity, limiting them to capturing excitatory interactions. 
To incorporate both excitatory and inhibitory effects, many studies~\citep{gerhard2017stability,linderman2016bayesian,mei2017neural} have proposed nonlinear Hawkes process that allows the influence functions to be negative. 
This study adopts the nonlinear Hawkes process proposed by \cite{zhou2022efficient} which is defined as: 
\begin{equation*}
\lambda^*(t) = \bar{\lambda} \sigma(h(t)), \quad h(t) = \mu + \sum\limits_{t_i < t} \phi(t - t_i), 
\label{intens}
\end{equation*}
where $\bar{\lambda}>0$ is the intensity upperbound, $\sigma(\cdot)$ denotes the sigmoid function, $\mu \in \mathbb{R}$ is the baseline activation and $\phi(\cdot)\in \mathbb{R}$ is the influence function. 
Due to the presence of the sigmoid function $\sigma(\cdot)$, both $\mu$ and $\phi(\cdot)$ can be negative, allowing it to capture inhibitory effects. 
We choose the sigmoid as the link function due to its compatibility with the subsequent data augmentation technique. 

For a flexible influence function, we model $\phi(\cdot)$ as a linear combination of multiple basis functions: 
\begin{equation*}
\phi(\cdot) = \sum\limits_{b=1}^B w_{b} \tilde{\phi}_b(\cdot),
\end{equation*}
where $\tilde{\phi}_b(\cdot)$ is the $b$-th basis function and $w_{b}\in \mathbb{R}$ is the mixing weight. 
Following \cite{zhou2022efficient}, we select the scaled and shifted beta densities with support $[0,T_\phi]$ as basis functions. 
We define the basis function with bounded support $[0,T_\phi]$ rather than unbounded support $[0,\infty]$ to assume that events occurring too early do not influence the current time. This choice ensures a more efficient computation of $\mathbf{\Phi}(t)$ afterward (details are provided in complexity analysis). 
Consequently, the formulation of $h(t)$ can be expressed in vector form: 
\begin{equation*}
h(t) = \mu + \sum_{t_i<t} \phi(t-t_i) = \mu + \sum\limits_{t_i<t} \sum_{b=1}^B w_{b} \tilde{\phi}_b(t-t_i) = \mu + \sum_{b=1}^B w_{b} \sum_{t_i<t} \tilde{\phi}_b(t-t_i) = \mathbf{w}^\top\bm{\Phi}(t), 
\end{equation*}
where $\mathbf{w}=[\mu,w_{1},\ldots,w_{B}]^\top$, $\bm{\Phi}(t)=[1,\Phi_{1}(t),\ldots,\Phi_{B}(t)]^\top$ and $\Phi_{b}(t)=\sum_{t_i<t}\tilde{\phi}_b(t-t_i)$ represents the cumulative impact of past events on $t$ through the $b$-th basis function. 
As a result, the probability density function of the proposed model is:
\begin{equation*}
    p(t_{1:N}|\mathbf{w},\bar{\lambda}) =\prod_{i=1}^N \bar{\lambda}\sigma(h(t_i))\exp{\left(-\int_0^T\bar{\lambda}\sigma(h(t))dt\right)}, 
\label{nonlikelihood}
\end{equation*}
where we assume $t_{1:N}$ are observed on $[0,T]$ and the model parameters are $\mathbf{w}$ and $\bar{\lambda}$. 

\subsection{Non-conjugate Bayesian CPD}
The above section outlines the Hawkes process without change points. In this section, we introduce the Hawkes process with change points and how the Bayesian two-step CPD is designed to detect these change points. 
In Bayesian two-step CPD, our goal is to identify the change point where the underlying dynamics of point process shift. The two steps in this method involve \emph{an estimation step} and \emph{a prediction step}. 
Let $t_{1:m}$ represent the sequence of timestamps that is generated by a Hawkes process with parameters $\bm{\theta}$, and we consider $\bm{\theta}$ may undergo changes at certain timestamps. 
We define, for the timestamp $t_m$, the nearest change point's index as $\tau_m \in \{1,\ldots,m\}$ and assume that the timestamps before and after the change point are mutually independent. 
Following this assumption, during \emph{the estimation step}, the Bayesian two-step CPD necessitates estimating the posterior of the model parameters based on $t_{\tau_m:m}$. 
To infer the posterior, we express the likelihood for the timestamps $t_{\tau_m:m}$ after the change point as: 
\begin{equation}
    p(t_{\tau_m:m}|\mathbf{w},\bar{\lambda}) =\prod_{i=\tau_m}^m \bar{\lambda}\sigma(h(t_i))\exp{\left(-\int_{t_{\tau_m}}^{t_m}\bar{\lambda}\sigma(h(t))dt\right)}. 
\label{likelihood}
\end{equation}
According to Bayes' theorem, the posterior of model parameters is expressed as: 
\begin{equation}
p(\mathbf{w}, \bar{\lambda}| t_{\tau_m:m})\propto p(t_{\tau_m:m}|\mathbf{w},\bar{\lambda})p(\mathbf{w})p(\bar{\lambda}), 
\label{eq3}
\end{equation}
where we choose the prior of $\mathbf{w}$ as Gaussian $p(\mathbf{w}) = \mathcal{N}(\mathbf{w}|\mathbf{0},\mathbf{K})$ and the prior of $\bar{\lambda}$ as an uninformative improper prior $p(\bar{\lambda})\propto 1/\bar{\lambda}$. 
We use a Gaussian prior on $\mathbf{w}$ because it is equivalent to an $L_2$ regularizer, which stabilizes parameter estimation when there is insufficient observed data. 

Then, in \emph{the prediction step}, we leverage the posterior of model parameters to compute the predictive distribution of the next timestamp as: 
\begin{equation}
p(t_{m+1}| t_{\tau_m:m})
=\iint p(t_{m+1}| t_{\tau_m:m},\mathbf{w}, \bar{\lambda})
p(\mathbf{w}, \bar{\lambda}| t_{\tau_m:m})
d\mathbf{w}d\bar{\lambda}. 
\label{preditive_cpd}
\end{equation}
This formula calculates the distribution of the next timestamp $t_{m+1}$ given the observed data points $t_{\tau_m:m}$. 
It is worth noting that this predictive distribution takes into account all possible model specifications, which is a key distinction between Bayesian and frequentist methods.

In implementation, solving \cref{eq3,preditive_cpd} is challenging. 
For \cref{eq3}, the non-conjugate nature between the point process likelihood and the prior prevents us from obtaining an analytical posterior. 
For \cref{preditive_cpd}, evaluating the integral 
is also intractable.
Therefore, we resort to sampling methods for approximation: 
\textbf{(1)} Use MCMC to obtain parameter samples from the posterior $\{\mathbf{w}^{(k)}, \bar{\lambda}^{(k)}\}_{k=1}^K\sim p(\mathbf{w}, \bar{\lambda}| t_{\tau_m:m})$. 
\textbf{(2)} Based on the sampled parameters, use the thinning algorithm~\citep{ogata1998space} to sample the next timestamp $\{t_{m+1}^{(k)}\sim p(t_{m+1}| t_{\tau_m:m},\mathbf{w}^{(k)}, \bar{\lambda}^{(k)})\}_{k=1}^K$. Create a confidence interval based on the samples of $\{t_{m+1}^{(k)}\}_{k=1}^K$. If the actual $t_{m+1}$ falls within this interval, we conclude that no change point has occurred. Conversely, we infer the presence of a change point.

\subsection{Conjugate Bayesian CPD}
For non-conjugate Bayesian CPD, the MCMC algorithm in step 1 often lacks analytical expressions, significantly impacting the timeliness of CPD due to its low computational efficiency. 
To address this issue, our CoBay-CPD adopts the data augmentation strategy, which augments the Hawkes process likelihood with auxiliary latent variables, enabling the augmented likelihood conditionally conjugate to the prior. 
Based on the conditionally conjugate model, we derive an analytical Gibbs sampler to effectively obtain posterior samples of parameters. 
Specifically, we incorporate P\'{o}lya-Gamma variables and marked Poisson processes. Similar derivations have been presented in~\cite{zhou2022efficient}; here we restate the key formulas for clarity. 

\subsubsection{Augmentation of P\'{o}lya-Gamma Variables}
The sigmoid function $\sigma(\cdot)$ can be represented in the form of a Gaussian scale mixture: 
\begin{equation*}
\sigma(z) = \int_0^{\infty}e^{f(\omega,z)}p_{\text{PG}}(\omega|1,0)d\omega,
\label{pygm}
\end{equation*}
where $f(\omega,z) = z/2 - z^2\omega/2-\log 2$ and $p_{\text{PG}}(\omega|1,0)$ denotes the P\'{o}lya-Gamma distribution with $\omega \in \mathbb{R}^{+}$. 
When substituting the above expression into the product term in \cref{likelihood}, the parameter $\mathbf{w}$ within the model takes on a Gaussian form. 

\subsubsection{Augmentation of Marked Poisson Process}
A marked Poisson process can be introduced to linearize the exponential integral term in \cref{likelihood}: 
\begin{equation*}
\exp{\left(-\int_{t_{\tau_{m}}}^{t_m}\bar{\lambda}\sigma(h(t))dt\right)}
= \mathbb{E}_{p_{\lambda}}\left[\prod_{(\omega, t) \in \Pi}e^{f(\omega, -h(t))}\right],
\end{equation*}
where $\Pi=\{(\omega_r, t_r)\}_{r=1}^R$ denotes a realization of a marked Poisson process on the interval $[t_{\tau_{m}},t_m]$, with its probability measure denoted as $p_{\lambda}$ and having an intensity $\lambda(t,\omega) = \bar{\lambda}p_{\text{PG}}(\omega|1,0)$. 
Notably, the key difference between our proposed change point model and the prior work by \cite{zhou2022efficient} lies in the fact that here, we focus on the interval with change points, $[t_{\tau_{m}},t_m]$, rather than the entire domain.

\subsubsection{Augmented Joint Distribution}
After introducing two sets of latent variables into \cref{likelihood}, we obtain the augmented likelihood: 
\begin{equation*}
p(t_{\tau_m:m}, \bm{\omega}, \Pi|\mathbf{w},\bar{\lambda})=\prod_{i=\tau_{m}}^m [\lambda(t_i, \omega_i)e^{f(\omega_i, h(t_i))}]p_{\lambda}(\Pi|\bar{\lambda})\prod_{(\omega,t) \in \Pi}e^{f(\omega,-h(t))}, 
\label{augmented likelihood}
\end{equation*}
where $\bm{\omega}$ is the vector of $\omega_i$ on each $t_i$ in $t_{\tau_m:m}$, $\lambda(t_i,\omega_i)=\overline{\lambda} p_{\text{PG}}(\omega_i|1,0)$. 
The parameter $\mathbf{w}$ in the augmented likelihood takes on a Gaussian form, making it conditionally conjugate to the Gaussian prior.
Combining the augmented likelihood with priors, we obtain the augmented joint distribution: 
\begin{equation*}
\begin{aligned}
&p(t_{\tau_m:m}, \bm{\omega}, \Pi, \mathbf{w}, \bar{\lambda}) = p(t_{\tau_m:m}, \bm{\omega}, \Pi|\mathbf{w},\bar{\lambda})p(\mathbf{w})p(\bar{\lambda}). 
\end{aligned}
\end{equation*}

\subsubsection{Gibbs Sampler}
Thanks to the conditional conjugacy of the augmented joint distribution, we can derive closed-form conditional densities for all variables, naturally leading to an analytical Gibbs sampler (derivation provided in \cref{app_gibbs}): 
\begin{subequations}
\begin{gather}
p(\bm{\omega}|t_{\tau_m:m},\mathbf{w}) = \prod_{i=\tau_m}^m p_{\text{PG}}(\omega_i | 1, h(t_i)), \label{omega} \\
\Lambda(t,\omega | t_{\tau_m:m}, \mathbf{w}, \bar{\lambda}) = \bar{\lambda}\sigma(-h(t))p_{\text{PG}}(\omega | 1,h(t)), \label{Lambda}\\
p(\bar{\lambda} | t_{\tau_m:m}, \Pi) = p_{\text{Ga}}(\bar{\lambda} | N_m + R, T_m), \label{lamb_bar}\\
p(\mathbf{w} | t_{\tau_m:m}, \bm{\omega}, \Pi) = \mathcal{N}(\mathbf{w}| \mathbf{m}, \bm{\Sigma}). \label{w}
\end{gather}
\label{gibbs_eq}
\end{subequations}
In \cref{lamb_bar}, $N_m = m-\tau_m + 1$, $R=|\Pi|$ is the number of points on the marked Poisson process, and $T_m = t_m - t_{\tau_m}$. In \cref{w}, $\bm{\Sigma} = [\bm{\Phi}\mathbf{D} \bm{\Phi}^{\top} + \mathbf{K}^{-1}]^{-1}$ where $\mathbf{D}$ is a diagonal matrix with $\{\omega_i\}_{i=\tau_m}^{m}$ in the first $m-\tau_m+1$ entries and $\{\omega_r\}_{r=1}^R$ in the following $R$ entries, and $\bm{\Phi} = [\{\bm{\Phi}(t_i)\}_{i=\tau_m}^m, \{\bm{\Phi}(t_r)\}_{r=1}^R]$; $\mathbf{m} = \bm{\Sigma}\bm{\Phi}\mathbf{v}$, where the first $m - \tau_m + 1$ entries of $\mathbf{v}$ are $1/2$, and the following $R$ entries are $-1/2$. 
Through iterative sampling using \cref{gibbs_eq}, we obtain a series of samples from the model parameter posterior. 
The pseudocode is provided in \cref{alg::Gibbs sampler}. 


\subsubsection{Algorithm, Hyperparameters and Complexity}
By employing the proposed Gibbs sampler for analytical posterior sampling of model parameters and subsequently using the thinning algorithm for prediction, we establish our two-step CoBay-CPD tailored for Hawkes process. The detailed procedure is outlined in \cref{alg::conjugate Bayesian online change point detection}. 

CoBay-CPD's hyperparameters, including covariance $\mathbf{K}$ in the Gaussian prior, confidence intervals, and basis functions, impact its performance. 
In experiments, we assume $\mathbf{K}=\sigma^2\mathbf{I}$.
Oversized $\sigma^2$ weakens the prior, causing unstable parameter estimation and oversensitive change point detection, while undersized values result in sluggish detection. 
Similarly, narrow confidence intervals oversensitize, and wide intervals slow detection. 
Balancing accuracy and efficiency, more basis functions enhance prediction accuracy but challenge computational efficiency. 
In experiments, we select all hyperparameters through cross-validation. 

Assuming the length of the entire sequence is $N$, the average length of $t_{\tau_m:m}$ is $M$, the average length of the latent marked Poisson process is $R$, the average number of points within the interval of $T_{\phi}$ is $N_{\phi}$, and the number of Gibbs iterations is $L$, the computation complexity of CoBay-CPD is $\mathcal{O}(N(MN_{\phi}B+LRN_{\phi}B+LC_{\text{TH}}+L(M+R)(B+1)^2+L(B+1)^3))$, where $C_{\text{TH}}$ represents the complexity of the thinning algorithm.
The detailed analysis is provided in \cref{Analysis of the complexity}. 


\section{Experiments}
We evaluate the performance of CoBay-CPD on both synthetic and real-world datasets. 
For the synthetic data, our aim is to validate the capability of CoBay-CPD in accurately recovering the ground-truth parameters and change points. 
For the real-world data, we compare CoBay-CPD against several baseline methods to determine whether our approach exhibits superior CPD performance.

\subsection{Baselines}
We conduct a comparison between CoBay-CPD and several Bayesian change point detection (BCPD) methods which are designed to address the non-conjugate challenge for Hawkes process: 
(1) \textbf{SMCPD}~\citep{doucet2009tutorial} combines BCPD and sequential Monte Carlo (SMC). Similar to our approach, it is a sampling-based method to address the non-conjugate inference in the BCPD framework. 
(2) \textbf{SVCPD}~\citep{detommaso2019stein} similarly combines BCPD and Stein variational inference to address the non-conjugate inference in the BCPD framework. Differently, this method infers the posterior of model parameters by the variational technique. 
(3) \textbf{SVCPD+Inhibition} is an extension of SVCPD that incorporates a nonlinear Hawkes process with inhibitory effects. This baseline is designed because the original SVCPD only considered a linear Hawkes process. 

\subsection{Metrics}
We use four metrics to assess the performance of all methods. 
(1) \textbf{False Negative Rate (FNR)} quantifies the probability of a change point being incorrectly identified as not a change point, calculated as $1 - \frac{\text{True Positives}}{\text{True Positives} + \text{False Negatives}}$. False negative cases are critical in many applications, as they indicate that certain crucial changes fail to receive attention. 
(2) \textbf{False Positive Rate (FPR)} quantifies the probability of a stable point being incorrectly identified as a change point, calculated as $1 - \frac{\text{True Negatives}}{\text{False Positives} + \text{True Negatives}}$. Minimizing false positive cases is also important because frequent false alarms will waste resources. 
(3) \textbf{Mean Square Error (MSE)}
measures the distance between the predicted next timestamp (the average of $t^{(k)}$) and the actual timestamp, calculated as $\frac{1}{n} \sum_{i=1}^{n} (\bar{t}^{(k)}_i - t_{i})^2$. This metric evaluates how accurately the model predicts the next data point. 
(4) \textbf{Running Time (RT)} measures the efficiency of the method by its runtime. 
 
\subsection{Synthetic Data}
We validate the efficacy of CoBay-CPD using a synthetic dataset. Our goal is to verify whether CoBay-CPD can accurately recover the ground-truth parameters and change points. 

\paragraph{Datasets}
The synthetic data is created by concatenating three segments of Hawkes process data, each characterized by different parameters. Specifically, all three segments of Hawkes processes adhere to the model configuration outlined in \cref{hawkes_inhibition}, with their influence functions assumed to be a mixture of multiple beta densities. Three segments employ identical basis functions and mixing weights. 
However, they have different intensity upperbounds, namely $\bar{\lambda}_1 = 5$, $\bar{\lambda}_2 = 10$ and $\bar{\lambda}_3 = 3$. We utilize the thinning algorithm to simulate data for three Hawkes processes, and concatenate them to form the synthetic data. The two change points are located at the $43$-rd and $136$-th points, indicated by grey lines in \cref{fig1b} in \cref{Synthetic Result Presentation}. 
Further details can be found in \cref{app_synthetic_data}. 

\begin{table}[t]
\caption{The FNR, FPR, MSE and RT of CoBay-CPD and other baselines on the synthetic dataset.}
\label{syn_table}
\begin{center}
\scalebox{0.85}{
\begin{tabular}{ccccc}
\toprule
Model & FNR($\downarrow$) & FPR(\% $\downarrow$) & MSE($\downarrow$) & RT(minute $\downarrow$) \\
\midrule
SMCPD  & 0.38 $\pm$ 0.41 &  0.76 $\pm$ 0.26  & 0.07 $\pm$ 0.01 & 5.50 $\pm$ 0.31 \\
SVCPD  & 0.50 $\pm$ 0.35 & 0.76 $\pm$ 0.26 & 0.06 $\pm$ 0.00 & 7.78 $\pm$ 0.01 \\
SVCPD+Inhi & 0.33 $\pm$ 0.24 & 0.60 $\pm$ 0.00  &  0.16 $\pm$ 0.01 &  23.09 $\pm$ 0.60 \\
CoBay-CPD & \bf{0.13 $\pm$ 0.22} & \bf{0.46 $\pm$ 0.26} & \bf{0.05 $\pm$ 0.00}  &  \bf{4.62 $\pm$ 0.10} \\
\bottomrule
\end{tabular}
}
\end{center}
\vspace{-0.1in}
\end{table}

\paragraph{Results}
We evaluate the performance of change point detection using CoBay-CPD and other baseline models on the synthetic dataset.
For CoBay-CPD, we adopt a prior distribution $p(\mathbf{w}) = \mathcal{N}(\mathbf{w}|\mathbf{0},\mathbf{K})$, where $\mathbf{K}=0.5\mathbf{I}$. 
The detection outcomes are presented in \cref{fig1b} in \cref{Synthetic Result Presentation}. The change points identified by CoBay-CPD are ${44, 136}$, while SMCPD detects ${96,136}$, SVCPD detects ${44,96}$ and SVCPD+Inhibition detects ${51,136}$. This discrepancy suggests that CoBay-CPD achieves more accurate change point detection. 
Furthermore, the estimated parameter $\bar{\lambda}$ from CoBay-CPD for the synthetic data is depicted in \cref{fig1d} in \cref{Synthetic Result Presentation}. 
The estimated parameter $\bar{\lambda}$ closely aligns with the ground truth, demonstrating the accuracy of parameter estimation by CoBay-CPD. 
Notably, there are prominent changes around the change points in \cref{fig1d}. This phenomenon arises due to the initiation of a new Hawkes process with distinct parameters at the occurrence of a change point. 
The challenge of accurately estimating parameters with limited data in such scenarios is alleviated by the Bayesian framework. In similar situations, frequentist methods tend to perform poorly. 

We also compare CoBay-CPD against baseline methods in terms of FNR, FPR, MSE and RT. The results are presented in \cref{syn_table}. As anticipated, CoBay-CPD outperforms the alternatives. This superiority can be attributed to CoBay-CPD's utilization of a nonlinear Hawkes process model, which encompasses both excitation and inhibition effects. In contrast, SMCPD and SVCPD employ a simpler linear Hawkes process model, constraining their expressive power. 
Additionally, CoBay-CPD employs Gibbs sampler to accurately characterize the parameter posterior, whereas both SVCPD and SVCPD+Inhibition utilize variational-based methods to approximate the parameter posterior. 
As a result, their change point detection accuracy is compromised.

\subsection{Real-world Data}
In this section, we conduct a comparison between CoBay-CPD and baselines on two real datasets. 

\paragraph{Datasets}
We analyze two datasets from the domains of network security and transportation, with specific details provided below. 
More comprehensive information regarding data preprocessing can be found in \cref{app_real_data}.
(1) \textbf{WannaCry Cyber Attack}\footnote{https://www.malware-traffic-analysis.net/2017/05/18/index2.html}~\citep{detommaso2019stein}: The WannaCry virus infected more than 200,000 computers around the world in 2017 and received much attention. The WannaCry Cyber Attack data contains 208 traffic logs information observations. Each observation contains the relevant timestamp. 
(2) \textbf{NYC Vehicle Collisions}\footnote{https://data.cityofnewyork.us/Public-Safety/NYPD-Motor-Vehicle-Collisions/h9gi-nx95}~\citep{zhou2020fast}: The New York City vehicle collision dataset comprises approximately 1.05 million vehicle collision records, each containing information about the time and location of the collision. For our experiments, we select the records from Oct.14th, 2017. 

The real-world data, unlike synthetic data, does not have ground-truth change points. 
Therefore, we use the points where timestamps surge as the ground-truth change points in the WannaCry dataset,
and utilize the reported change points from~\cite{zhou2020fast} as the ground-truth change points in the NYC dataset. 

\begin{table*}[t]
\caption{The FNR, FPR, MSE and RT of CoBay-CPD and other baselines on real-world datasets.}
\label{tll_acc_result}
\begin{center}
\scalebox{0.70}{
\begin{tabular}{c|cccc|cccc}
\toprule
\multirow{2}{*}{Model} & \multicolumn{4}{c|}{WannaCry} & \multicolumn{4}{c}{NYC Vehicle Collisions}\\
\cmidrule{2-9}
& FNR($\downarrow$) & FPR($\downarrow$) & MSE($\times$10$^2 \downarrow$)& RT(minute $\downarrow$) & FNR($\downarrow$) & FPR(\% $\downarrow$)& MSE($\downarrow$) &RT(minute $\downarrow$)\\
\midrule
SMCPD & 0.38 $\pm$ 0.06 & 0.02 $\pm$ 0.01 & 3.59 $\pm$ 0.08 & 11.65 $\pm$ 0.07 &0.56 $\pm$ 0.16 & 2.46 $\pm$ 0.55 & 0.02 $\pm$ 0.00 & 24.67 $\pm$ 0.26\\
SVCPD & 0.34 $\pm$ 0.12 & 0.01 $\pm$ 0.01 & 3.47 $\pm$ 0.06 & 9.72 $\pm$ 0.06 & 0.58 $\pm$ 0.36 & 1.00 $\pm$ 0.43 & 0.02 $\pm$ 0.00 & 19.30 $\pm$ 0.09\\
SVCPD+Inhi & 0.54 $\pm$ 0.09 & \bf{0.00 $\pm$ 0.00} & 3.54 $\pm$ 0.06 & 29.76 $\pm$ 2.54 & 0.22 $\pm$ 0.16 & 1.55 $\pm$ 0.36 & 0.17 $\pm$ 0.01 & 64.47 $\pm$ 1.36\\
CoBay-CPD  & \bf{0.21 $\pm$ 0.04} & 0.05 $\pm$ 0.02 & \bf{3.42 $\pm$ 0.00 }& \bf{6.24 $\pm$ 0.49} & \bf{0.13 $\pm$ 0.16} & \bf{0.89 $\pm$ 0.16} & \bf{0.01 $\pm$ 0.00} & \bf{8.70 $\pm$ 0.26}\\
\bottomrule
\end{tabular}}
\end{center}
\end{table*}
\begin{table*}[t]
\caption{Ablation study. The FNR, FPR, MSE and RT of CoBay-CPD with different hyperparameters. 
}
\label{Ablation Study}
\begin{center}
\scalebox{0.65}{
\begin{tabular}{c|ccc|ccc|ccc}
\toprule
\multirow{2}{*}{Metric} & \multicolumn{3}{c|}{Number of Basis Functions}                         & \multicolumn{3}{c|}{Confidence Interval}                                                            & \multicolumn{3}{c}{Prior Covariance} \\
\cmidrule{2-10}
                  & $1$           & $2$           & $3$           & $95\%$ & $90\%$ & $85\%$ & $\sigma^2=0.01$       &   $\sigma^2=0.5$       & $\sigma^2=10$ \\ \midrule
FNR($\downarrow$)               & 0.38 $\pm$ 0.41 & 0.38 $\pm$ 0.22 & \bf{0.13 $\pm$ 0.22} & 0.50 $\pm$ 0.00                  & \bf{0.13 $\pm$ 0.22}              & 0.25 $\pm$ 0.25                  & \bf{0.13 $\pm$ 0.22} & \bf{0.13 $\pm$ 0.22} & 0.50 $\pm$ 0.00 \\
FPR(\% $\downarrow$)           & 1.07 $\pm$ 0.50 & 0.91 $\pm$ 0.30 & \bf{0.61 $\pm$ 0.00} & \bf{0.46 $\pm$ 0.26}                  & \bf{0.46 $\pm$ 0.26}              & 1.83 $\pm$ 0.43                  & 0.76 $\pm$ 0.26 & \bf{0.46 $\pm$ 0.26} & 0.91 $\pm$ 0.30 \\
MSE($\downarrow$)               & \bf{0.05 $\pm$ 0.00} & \bf{0.05 $\pm$ 0.00} & \bf{0.05 $\pm$ 0.00} & \bf{0.04 $\pm$ 0.00}                  & 0.05 $\pm$ 0.00              & \bf{0.04 $\pm$ 0.00}                  & \bf{0.04 $\pm$ 0.00} & 0.05 $\pm$ 0.00 & 0.05 $\pm$ 0.01 \\
RT(minute $\downarrow$)               & \bf{1.57 $\pm$ 0.03} & 2.61 $\pm$ 0.08 & 3.62 $\pm$ 0.10 & 5.03 $\pm$ 0.02                  & 4.62 $\pm$ 0.10              & \bf{4.50 $\pm$ 0.11}                  & 4.74 $\pm$ 0.02 & 4.62 $\pm$ 0.10 & \bf{4.41 $\pm$ 0.10} \\
\bottomrule
\end{tabular}
}
\end{center}
\vspace{-0.1in}
\end{table*}

\paragraph{Results}
\Cref{fig2b,fig2c,fig2d,fig2e} in \cref{app_complete_diagram} display the change point detection outcomes of different methods applied to WannaCry data. 
It is clear that CoBay-CPD exhibits the most favorable detection performance. 
The change points identified by our method are consistent with the actual change points. The SMCPD, SVCPD and SVCPD+Inhibition detect a relatively limited number of change points, resulting in missed change points. 
\Cref{tll_acc_result} presents various metrics of four methods for change point detection in the WannaCry data. 
Clearly, because SMCPD, SVCPD, and SVCPD+Inhibition detect too few change points, their FNR is high and FPR is low. In contrast, CoBay-CPD exhibits the lowest FNR, a reasonably balanced FPR, the smallest MSE, and requires the least runtime. 

\Cref{fig3b,fig3c,fig3d,fig3e} in \cref{app_complete_diagram} show the change point detection outcomes of four methods for the NYC data. 
Notably, SVCPD detects fewer change points, while SMCPD identify an excessive number. 
The change points detected by CoBay-CPD are 43, 110, 160, 194, 284, 338, 398, corresponding to the times 2:30, 9:00, 12:00, 13:10, 16:00, 17:55, 20:00. These timestamps coincide with peak traffic hours on workdays. 
\Cref{tll_acc_result} presents various metrics of four methods for change point detection in the NYC data. 
Consistently, SMCPD exhibits high FNR and FPR. The high FPR is due to an excessive number of change points detected by SMCPD, while the high FNR is due to the inaccurate detection of numerous change points by SMCPD.
Whereas SVCPD shows a high FNR and low FPR due to detecting too few change points. SVCPD+Inhibition achieves a relatively balanced FNR and FPR, indicating the beneficial impact of employing a nonlinear Hawkes process. 
CoBay-CPD demonstrates superior accuracy and efficiency, with the lowest FNR, FPR, MSE, and RT in change point detection compared to all baseline models.

\subsection{Ablation Study}
In this section, we conduct hyperparameter analysis and stress tests of CoBay-CPD on synthetic data. 

\paragraph{Number of Basis Functions}
The number of basis functions impacts the expressiveness of the Hawkes process, influencing the model's detection performance. We assess the model's detection performance across varying numbers of basis functions, from 1 to 3, as shown in the \cref{Ablation Study}. 
In experiment, we set the other hyperparameters as: 90\% confidence interval and $\mathbf{K}=0.5\mathbf{I}$. 
Observably, as the number of basis functions increases, FNR and FPR decrease, indicating enhanced detection accuracy, while RT increases, indicating increased computational burden. 

\paragraph{Confidence Interval}
We try 3 different confidence intervals: 95\%, 90\%, and 85\% for the next timestamp, as shown in \cref{Ablation Study}. 
In the experiment, we choose 4 basis functions and $\mathbf{K}=0.5\mathbf{I}$.
A wider confidence interval results in fewer detected change points, leading to a larger FNR and a smaller FPR. Conversely, a narrower confidence interval leads to the detection of more change points, resulting in numerous incorrect change points. Consequently, the FPR increases significantly, while the FNR also shows a slight rise. 
So the compromise, 90\% confidence intervals, is the best.

\paragraph{Prior Covariance}
The Gaussian prior covariance $\mathbf{K}=\sigma^2\mathbf{I}$ also has a large impact on the detection results. 
In this experiment, we choose 4 basis functions and 90\% confidence interval. 
If $\sigma^2$ is too large, FNR and FPR will increase. This is because the prior is too loose, causing the posterior samples of model parameters to spread excessively and fail to concentrate around the true values. 
On the contrary, when $\sigma^2$ is too small, the posterior samples of model parameters are too concentrated in a certain position that may be a wrong value, resulting in a larger FPR, as shown in \cref{Ablation Study}. 

\paragraph{Stress Tests}
The stress tests assess how well a model performs under difficult or extreme conditions. We conduct three stress tests experiments: one involving the number of change points (more indicating greater difficulty), another focusing on the difference between adjacent $\bar{\lambda}$'s, $\Delta \bar{\lambda}$ (smaller indicating greater difficulty), and the third examining the closeness between adjacent change points, $\Delta t$ (smaller indicating greater difficulty). The results are shown in \cref{stree_test}. 
(1) Experiments with different numbers of change points reveal consistent performance across varying numbers. 
(2) Regarding $\Delta \bar{\lambda}$, our model effectively detects change points even when $\Delta \bar{\lambda}$ is small (e.g., $\Delta \bar{\lambda} = 1$). However, excessively small $\Delta \bar{\lambda}$ values lead to decreased performance, as the parameters on both sides of the change point become too similar to distinguish. 
(3) Regarding $\Delta t$, the model maintains good even when two change points are close, although performance slightly declines. 
More experimental details can be found in \cref{ST}. 


\begin{table}[t]
\caption{The results of stress tests. We conduct experiments to verify the performance change of CoBay-CPD w.r.t. the number of change points, the difference between adjacent $\bar\lambda$'s, and the closeness between two change points.}
\label{stree_test}
\begin{center}
\scalebox{0.67}{
\begin{tabular}{c|ccc|ccc|ccc}
\toprule
\multirow{2}{*}{Metric} & \multicolumn{3}{c|}{Number of Change Points}        & \multicolumn{3}{c|}{ $\Delta \bar{\lambda}$}      & \multicolumn{3}{c}{$\Delta t$ }\\ 
\cmidrule{2-10}
                        & 1               & 2               & 3               & 0.1             & 1               & 5               & 5                   & 10                  & 15                 \\
\midrule
FNR( $\downarrow$)                  & \bf{0.00 $\pm$ 0.00} & 0.13 $\pm$ 0.22 & 0.11 $\pm$ 0.14 & 1.00 $\pm$ 0.00 & 0.25 $\pm$ 0.43 & \bf{0.00 $\pm$ 0.00} & 0.33 $\pm$ 0.24     & \bf{0.00 $\pm$ 0.00}     & \bf{0.00 $\pm$ 0.00}               \\
FPR(\% $\downarrow$)                     & 0.43 $\pm$ 0.60 & 0.46 $\pm$ 0.26 & \bf{0.31 $\pm$ 0.50} & 1.61 $\pm$ 0.57 & \bf{0.35 $\pm$ 0.60} & 0.43 $\pm$ 0.60 & 1.00 $\pm$ 0.70     & 0.93 $\pm$ 0.65     & 0.31 $\pm$ 0.54                \\
MSE( $\downarrow$)                      & \bf{0.04 $\pm$ 0.00} & 0.05 $\pm$ 0.00 & 0.07 $\pm$ 0.01 & \bf{0.02 $\pm$ 0.00} & 0.03 $\pm$ 0.00 & 0.04 $\pm$ 0.00 & \bf{0.01 $\pm$ 0.00}     & 0.02 $\pm$ 0.00     & 0.02 $\pm$ 0.00              \\ 
\bottomrule
\end{tabular}
}
\end{center}
\vspace{-0.1in}
\end{table}

\section{Limitations and Broader Impacts}

Although our proposed CoBay-CPD method
offers an efficient and accurate solution to the change point detection problem in Hawkes process, it still has some limitations. For instance, extending CoBay-CPD to multivariate Hawkes processes remains challenging because the current method requires change points to occur at specific event locations. However, in multivariate Hawkes processes, a change point in one dimension (an event location) does not necessarily correspond to event locations in other dimensions. This challenge needs to be addressed further in future research. 

The introduction of CoBay-CPD for Hawkes process holds promise for both positive and negative social impacts. 
This method effectively addresses the non-conjugate inference challenge, improving the efficiency and accuracy of change point detection across various fields. 
However, as automated change point detection methods become more accurate and efficient, there is a risk of overreliance on these systems without proper validation or human oversight. This could lead to erroneous decisions or missed opportunities for critical interventions. 

\section{Conclusions}
In summary, this work introduces a novel conjugate Bayesian two-step change point detection method for Hawkes process, which effectively addresses the non-conjugate inference challenge. 
Leveraging data augmentation, we transform the non-conjugate inference problem to a conditionally conjugate one, enabling the development of an analytical Gibbs sampler for efficient parameter posterior sampling. 
Our proposed approach surpasses existing methods, showcasing superior accuracy and efficiency in detecting change points. 
The contributions of this research hold great potential for advancing event-driven time series analysis and change point detection across various applications.

\begin{ack}
This work was supported by NSFC Projects (Nos. 62106121, 72171229), the MOE Project of Key Research Institute of Humanities and Social Sciences (22JJD110001), the Big Data and Responsible Artificial Intelligence for National Governance, Renmin University of China, the fundamental research funds for the central universities, and the research funds of Renmin University of China (24XNKJ13). 
\end{ack}

\bibliographystyle{icml2022}
\bibliography{ijcai24}


\newpage

\appendix

\section{Derivation of CoBay-CPD}
\label{app_gibbs}

In this section we provide proof of data augmentation and Gibbs sampler, respectively.

\subsection{Data Augmentation}
Focus on the interval with change points $[t_{\tau_{m}},t_m]$, the probability density (likelihood) of CoBay-CPD can be presented as:
\begin{equation*}
    p(\{t_i\}_{i=\tau_m}^m|\mathbf{w},\bar{\lambda}) =\prod_{i=\tau_m}^m \bar{\lambda}\sigma(h(t_i))\exp{\left(-\int_{t_{\tau_m}}^{t_m}\bar{\lambda}\sigma(h(t))dt\right)}. 
\end{equation*}

Substitute the augmentation of P\'{o}lya-Gamma Variables $\sigma(z) = \int_0^{\infty}e^{f(\omega,z)}p_{\text{PG}}(\omega|1,0)d\omega$ and the sigmoid symmetry property $\sigma(z) = 1- \sigma(-z)$ to the above equation, we can obtain:
\begin{equation*}
\exp{\left(-\int_{t_{\tau_m}}^{t_m}\bar{\lambda}\sigma(h(t))dt\right)} = \exp{\left(-\int_{t_{\tau_m}}^{t_m}\int_0^{\infty}(1 -e^{ f(\omega,-h(t))})\bar{\lambda}p_{\text{PG}}(\omega|1,0)d\omega dt)\right)}. 
\end{equation*}

According to Campbell's theorem \citep{kingman1992poisson}, the exponential integral term can be rewritten as 
\begin{equation*}
\exp{\left(-\int_{t_{\tau_{m}}}^{t_m}\bar{\lambda}\sigma(h(t))dt\right)}
= \mathbb{E}_{p_{\lambda}}\left[\prod_{(\omega, t) \in \Pi}e^{f(\omega, -h(t))}\right],
\end{equation*}
where $\Pi={(\omega_r, t_r)}_{r=1}^R$ denotes a realization of a marked Poisson process on the interval $[t_{\tau_{m}},t_m]$, with its probability measure denoted as $p_{\lambda}$ and having an intensity $\lambda(\omega,t) = \bar{\lambda}p_{\text{PG}}(\omega|1,0)$. 

Therefore, the likelihood of CoBay-CPD can be rewritten as :
\begin{align*}
    p(\{t_i\}_{i=\tau_m}^m|\mathbf{w},\bar{\lambda}) &=\prod_{i=\tau_m}^m \bar{\lambda}\sigma(h(t_i))\exp{\left(-\int_{t_{\tau_m}}^{t_m}\bar{\lambda}\sigma(h(t))dt\right)} \\
    &=\prod_{i=\tau_m}^m \left(\int_0^{\infty}\bar{\lambda}e^{f(\omega_i,h(t_i)}p_{\text{PG}}(\omega_i|1,0)d\omega_i \right)\mathbb{E}_{p_{\lambda}}\left[\prod_{(\omega, t) \in \Pi}e^{f(\omega, -h(t))}\right] \\
    &=\int \int \prod_{i=\tau_{m}}^m [\lambda(t_i, \omega_i)e^{f(\omega_i, h(t_i))}]p_{\lambda}(\Pi|\bar{\lambda})\prod_{(\omega,t) \in \Pi}e^{f(\omega,-h(t))}d\bm{\omega} d\Pi,
\end{align*}
where $\bm{\omega}$ is the vector of $\omega_i$. It is straightforward to see the integrand is the augmented likelihood:
\begin{equation*}
p(\{t_i\}_{i=\tau_m}^m, \bm{\omega}, \Pi|\mathbf{w},\bar{\lambda})=
\prod_{i=\tau_{m}}^m [\lambda(t_i, \omega_i)e^{f(\omega_i, h(t_i))}]p_{\lambda}(\Pi|\bar{\lambda})\prod_{(\omega,t) \in \Pi}e^{f(\omega,-h(t))}.
\end{equation*}

\subsection{Gibbs Sampler}
Based on the augmented joint distribution \begin{equation*}
p(\{t_i\}_{i=\tau_m}^m, \bm{\omega}, \Pi, \mathbf{w}, \bar{\lambda} | \tau_m) =
p(\{t_i\}_{i=\tau_m}^m, \bm{\omega}, \Pi|\mathbf{w},\bar{\lambda},\tau_{m})p(\mathbf{w})p(\bar{\lambda}), 
\end{equation*}
we can derive the conditional densities of all variables in closed form. By sampling from these conditional densities iteratively, we construct an analytical Gibbs sampler.

\subsubsection{Derivation for $\omega$}
\begin{equation*}
    p(\bm{\omega}|\{t_i\}_{i=\tau_m}^m, \mathbf{w}) \propto \prod_{i=\tau_m}^m \left[ \lambda(t_i, \omega_i)e^{f(\omega_i, h(t_i))} \right]\prod_{(\omega,t) \in \Pi} e^{f(\omega,-h(t))},
\end{equation*}
 where $\prod\limits_{(\omega,t) \in \Pi} e^{f(\omega,-h(t))}$ is constant as $\Pi$ is given. In addition, combined  $f(\omega,z) = z/2 - z^2\omega/2-\log 2$ and $p_{\text{PG}}(\omega | b, 0) \cdot e^{-\frac{c^2 \omega}{2}} \propto p_{\text{PG}}(\omega | b, c)$ with the above derivation, we can deduce that,
\begin{align*}
    p(\bm{\omega}|\{t_i\}_{i=\tau_m}^m, \mathbf{w}) &\propto \prod_{i=\tau_m}^m \left[ \bar{\lambda} p_{\text{PG}}(\omega_i | 1, 0) e^{h(t_i)/2 - {h(t_i)}^2\omega/2-\log 2} \right] \propto \prod_{i=\tau_m}^m p_{\text{PG}}(\omega_i | 1, h(t_i)).
\end{align*}

\subsubsection{Derivation for $\Pi$}

The posterior of $\Pi$ is dependent on $\{t_i\}_{i=\tau_m}^m$, $\mathbf{w}$ and $\bar{\lambda}$,
\begin{equation*}
    p(\Pi | \{t_i\}_{i=\tau_m}^m, \mathbf{w}, \bar{\lambda}) = \frac{p_{\lambda}(\Pi| \bar{\lambda}) \prod_{(\omega,t) \in \Pi} e^{f(\omega,-h(t))}}{ \int p_{\lambda}(\Pi| \bar{\lambda}) \prod_{(\omega,t) \in \Pi} e^{f(\omega,-h(t))} d\Pi},
\end{equation*}
where Campbell's theorem can be applied to convert the denominator, the equation above can be transformed as
\begin{align*}
    &p(\Pi | \{t_i\}_{i=\tau_m}^m, \mathbf{w}, \bar{\lambda}) = \frac{p_{\lambda}(\Pi| \bar{\lambda}) \prod_{(\omega,t) \in \Pi} e^{f(\omega,-h(t))}}{\exp{\left(-\int_{t_{\tau_m}}^{t_m}\int_0^{\infty}(1 -e^{ f(\omega,-h(t))})\bar{\lambda}p_{\text{PG}}(\omega|1,0)d\omega dt)\right)}}  \\
    &= \prod_{(\omega,t) \in \Pi} \left(e^{ f(\omega,-h(t))})\bar{\lambda}p_{\text{PG}}(\omega|1,0)\right) \exp{\left(-\int_{t_{\tau_m}}^{t_m}\int_0^{\infty}e^{ f(\omega,-h(t))}\bar{\lambda}p_{\text{PG}}(\omega|1,0)d\omega dt)\right)}.
\end{align*}
The above posterior is in the likelihood form of a marked Poisson process with intensity
\begin{equation*}
    \Lambda(t,\omega | \{t_i\}_{i=\tau_m}^m, \mathbf{w}, \bar{\lambda}) = e^{ f(\omega,-h(t))})\bar{\lambda}p_{\text{PG}}(\omega|1,0) = \bar{\lambda}\sigma(-h(t))p_{\text{PG}}(\omega | 1,h(t)).
\end{equation*}
    
\subsubsection{Derivation for $\bar{\lambda}$}
\begin{align*}
    p(\bar{\lambda} | \{t_i\}_{i=\tau_m}^m, \Pi) &\propto \prod_{i=\tau_{m}}^m [\lambda(t_i, \omega_i)e^{f(\omega_i, h(t_i))}] p_{\lambda}(\Pi| \bar{\lambda})\cdot 1/\bar{\lambda} \\
    &\propto \prod_{i=\tau_{m}}^m [\bar{\lambda}p_{\text{PG}}(\omega | 1,h(t_i))] p_{\lambda}(\Pi| \bar{\lambda})\cdot 1/\bar{\lambda},
\end{align*}
where $p_{\text{PG}}(\omega | 1,h(t)) = \prod\limits_{(\omega,t) \in \Pi} \bar{\lambda}p_{\text{PG}}(\omega | 1,0)e^{-\int_{t_{\tau_m}}^{t_m} \int_{0}^{\infty} \bar{\lambda}p_{\text{PG}}(\omega | 1,0) d\omega dt}$. So the above equation are transformed as
\begin{equation*}
    p(\bar{\lambda} | \{t_i\}_{i=\tau_m}^m, \Pi) \propto \bar{\lambda}^{(m-\tau_m + 1 +|\Pi| - 1)} e^{-\int_{t_{\tau_m}}^{t_m}dt \cdot \bar{\lambda}}.
\end{equation*}
Let $N_m = m-\tau_m + 1$, $R=|\Pi|$ is the number of points on the marked Poisson process, and $T_m = t_m - t_{\tau_m}$, then we can get
\begin{equation*}
    p(\bar{\lambda} | \{t_i\}_{i=\tau_m}^m, \Pi) = p_{\text{Ga}}(\bar{\lambda} | N_m + R, T_m).
\end{equation*}

\subsubsection{Derivation for $\mathbf{w}$}

 For $\mathbf{w}$, we utilize the Gaussian prior $p(\mathbf{w}) = N(\mathbf{w}|\mathbf{0},\mathbf{K})$ where $\mathbf{K}$ is the prior covariance matrix.
 Then the derivation for $\mathbf{w}$ is as follows
\begin{align*}
    p(\mathbf{w} | \{t_i\}_{i=\tau_m}^m, \bm{\omega}, \Pi) &\propto \prod_{b=1}^{B+1} \frac{1}{\sqrt{2 \pi \sigma^2}} e^{-\frac{(w_b-\mu)^2}{2\sigma^2}} \prod_{i=\tau_m}^m e^{f(\omega_i,h(t_i))}  \prod_{(\omega,t) \in \Pi} e^{f(\omega,-h(t))},
\end{align*}
where $h(t) = \mu + \sum_{b=1}^B w_{b} \sum_{t_i<t} \tilde{\phi}_b(t-t_i)$ and $f(\omega,z) = z/2 - z^2\omega/2-\log 2$ and $p_{\text{PG}}(\omega | b, 0) \cdot e^{-\frac{c^2 \omega}{2}} \propto p_{\text{PG}}(\omega | b, c)$. Therefore, we obtain
\begin{equation*}
    p(\mathbf{w} | \{t_i\}_{i=\tau_m}^m, \bm{\omega}, \Pi) = N(\mathbf{w}| \mathbf{m}, \bm{\Sigma}),
\end{equation*}
where $\bm{\Sigma} = [\bm{\Phi}\mathbf{D} \bm{\Phi}^{\top} + \mathbf{K}^{-1}]^{-1}$, where $\mathbf{D}$ is a diagonal matrix with $\{\omega_i\}_{i=\tau_m}^{m}$ in the first $m-\tau_m+1$ entries and $\{\omega_r\}_{r=1}^R$ in the following $R$ entries, and $\bm{\Phi} = [\{\bm{\Phi}(t_i)\}_{i=\tau_m}^m, \{\bm{\Phi}(t_r)\}_{r=1}^R]$; $\mathbf{m} = \bm{\Sigma}\bm{\Phi}\mathbf{v}$, where the first $m - \tau_m + 1$ entries of $\mathbf{v}$ are $1/2$, and the following $R$ entries of $\mathbf{v}$ are $-1/2$.

\section{Algorithm}\label{Algorithm}
\subsection{Gibbs Sampler}
\label{alg::Gibbs sampler} 
\begin{algorithm}[!h] 
\renewcommand{\algorithmicrequire}{\textbf{Input:}}
\renewcommand{\algorithmicensure}{\textbf{Output:}}
\caption{Gibbs Sampler} 
\label{Gibbs sampler algo}
\begin{algorithmic}[1] 
\REQUIRE 
$t_{\tau_m:m}$, basis functions $\{\widetilde{\phi}_b(\cdot)\}_{b=1}^B$, covariance $\mathbf{K}$; 
\ENSURE 
Parameter posterior samples $\{\mathbf{w}^{(k)}, \bar{\lambda}^{(k)}\}_{k=1}^K$;
\FOR{iteration} 
    \STATE Sample $\bm{\omega}$ by \cref{omega};
    \STATE Sample $\Pi$ through thinning by \cref{Lambda};
    \STATE Sample $\bar{\lambda}$ by \cref{lamb_bar};
    \STATE Sample $\mathbf{w}$ by \cref{w}. 
    \ENDFOR
\end{algorithmic} 
\end{algorithm}

\subsection{CoBay-CPD}
\label{alg::conjugate Bayesian online change point detection}
\begin{algorithm}[!h] 
	\renewcommand{\algorithmicrequire}{\textbf{Input:}}
	\renewcommand{\algorithmicensure}{\textbf{Output:}}
	\caption{$(m+1)$-th round of CoBay-CPD} 
 \label{alg::conjugate Bayesian}
	\begin{algorithmic}[1] 
    \REQUIRE $t_{\tau_m:m+1}$, basis functions $\{\widetilde{\phi}_b(\cdot)\}_{b=1}^B$, covariance $\mathbf{K}$; 
    \ENSURE Change point at the current position or not; 
    \STATE Sample $\{\mathbf{w}^{(k)}, \bar{\lambda}^{(k)}\}_{k=1}^K$ by \cref{alg::Gibbs sampler}; 
    \STATE Sample $\{t_{m+1}^{(k)}\sim p(t_{m+1}| t_{\tau_m:m},\mathbf{w}^{(k)}, \bar{\lambda}^{(k)})\}_{k=1}^K$ and create a confidence interval of $\{t_{m+1}^{(k)}\}$, e.g., the 5\% and 95\% quantiles denoted as $t_{m+1}^l$ and $t_{m+1}^r$. If the actual $t_{m+1}$ falls within the interval $[t_{m+1}^l,t_{m+1}^r]$, classify it as not a change point; otherwise, a change point. 
\end{algorithmic} 
\end{algorithm}

\section{Analysis of Complexity}\label{Analysis of the complexity}
Assuming the length of the entire sequence is $N$, the average length of $t_{\tau_m:m}$ is $M$, the average length of the latent marked Poisson process is $R$, the average number of points within the interval of $T_{\phi}$ is $N_{\phi}$, and the number of Gibbs iterations is $L$, the computation complexity of CoBay-CPD is $\mathcal{O}(N(MN_{\phi}B+LRN_{\phi}B+LC_{\text{TH}}+L(M+R)(B+1)^2+L(B+1)^3))$, where $C_{\text{TH}}$ represents the complexity of the thinning algorithm in marked poisson process, not in the prediction step. This is because thinning algorithm in prediction step only samples one point at a time, which is very fast, so its computational complexity can be ignored.
We ignore the complexities of other sampling operations since they are fast. 
The first term corresponds to the precomputation of $\bm{\Phi}(t)$ on $t_{\tau_m:m}$, the second term to the computation of $\bm{\Phi}(t)$ on $\Pi$, the third term to the sampling of the marked Poisson process, the fourth and fifth terms to the computation of mean and covariance. 
By limiting the maximum length of $t_{\tau_m:m}$ and $T_{\phi}$, we can reduce the values of $M$, $N_\phi$ and $R$, thereby accelerating the computation of $\mathbf{\Phi}(t)$. 
Moreover, as the length of $t_{\tau_m:m}$ decreases, $C_{\text{TH}}$ will decrease as well. 

\section{Synthetic Data Experiment}
\label{app_synthetic_data}
\subsection{Data Processing}
We generate a synthetic data concatenated by three segments of Hawkes process data. In these three segments of Hawkes process data, we assume $4$ scaled beta densities: $\tilde{\phi}_{1,2,3,4}= \text{Beta}(\tilde{\alpha}=50, \tilde{\beta} = 50, \text{scale} = 6, \text{shift} = \{-2,-1,0,1\})$ as the basis functions with support $[0, T_{\phi}=6]$ and $\mu = 0$ as the baseline activation. However, they have different intensity upperbounds, namely $\bar{\lambda}_1 = 5$, $\bar{\lambda}_2 = 10$ and $\bar{\lambda}_3 = 3$. We use the thinning algorithm to generate a sequence according to the intensity function specified above.

\subsection{Result Presentation}
\label{Synthetic Result Presentation}
\begin{figure}[!h]
\centering
\begin{minipage}[b]{0.40\textwidth}
\centering
\includegraphics[width=\linewidth]{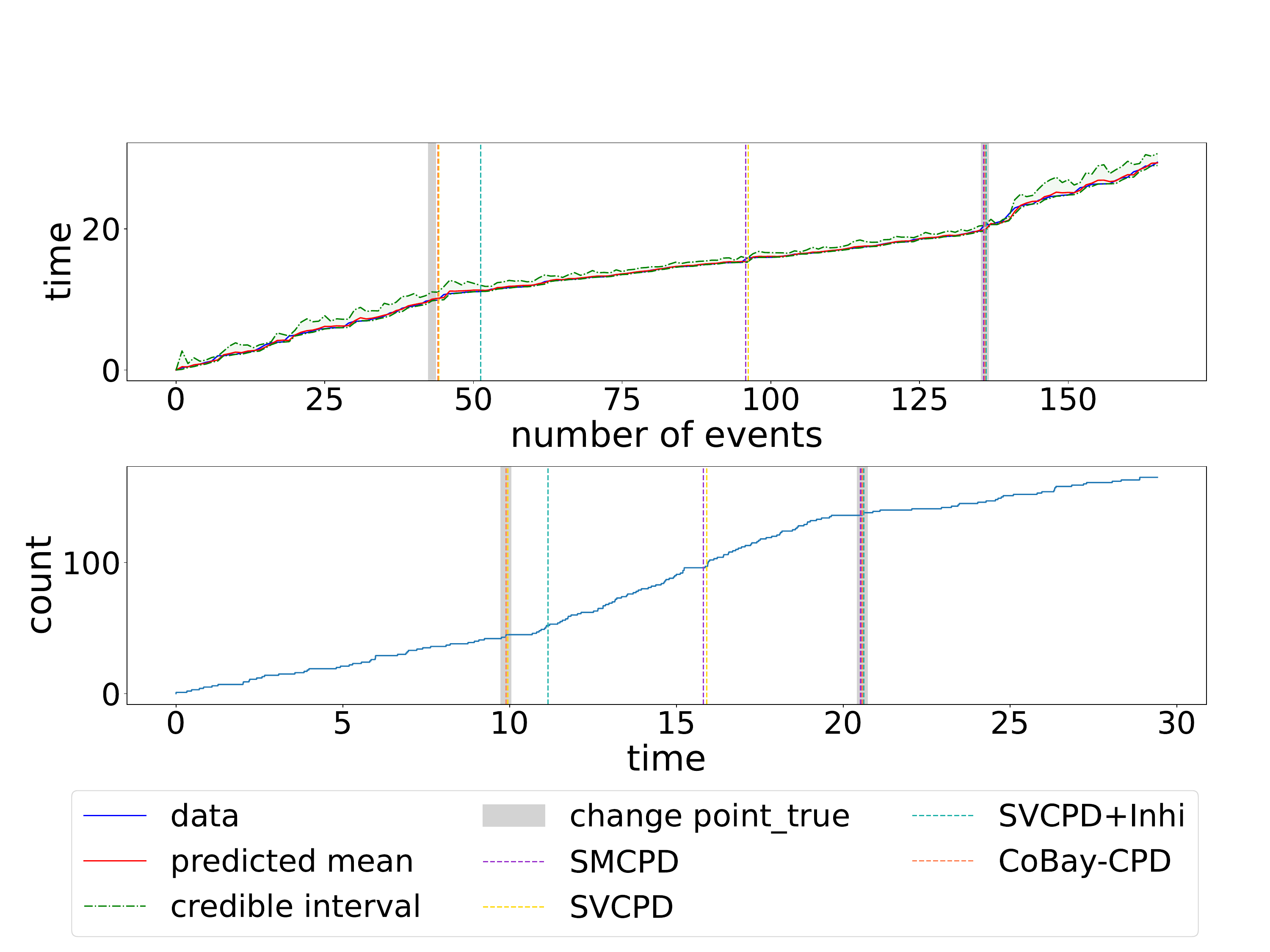}
\subcaption[]{Change Point Detection}
\label{fig1b}
\end{minipage}
\begin{minipage}[b]{0.30\textwidth}
\centering
\includegraphics[width=\linewidth]{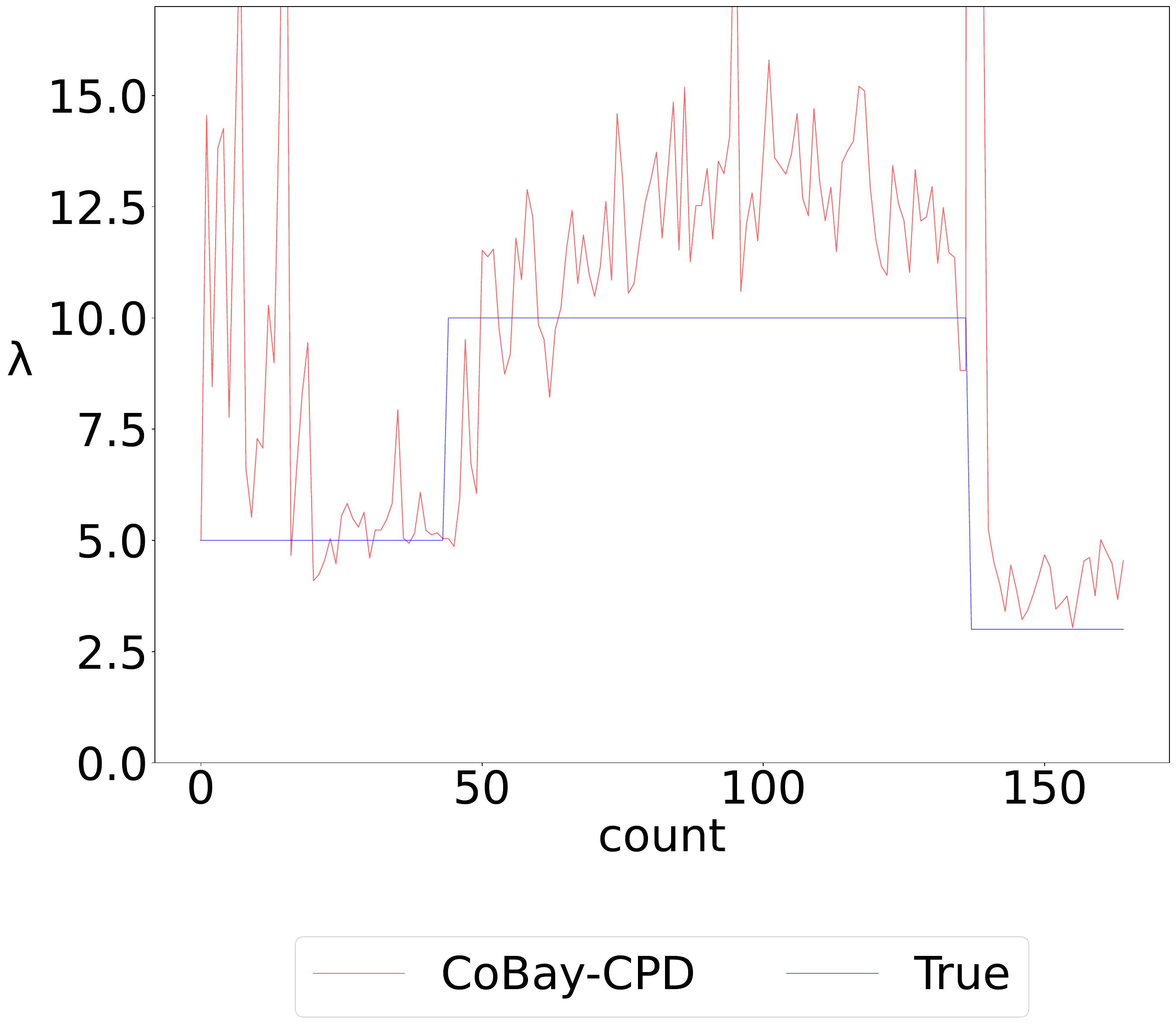}
\subcaption[]{Estimated $\bar{\lambda}$}
\label{fig1d}
\end{minipage}
\centering
   \begin{minipage}[b]{0.32\textwidth}
     \centering
     \includegraphics[width=\linewidth]{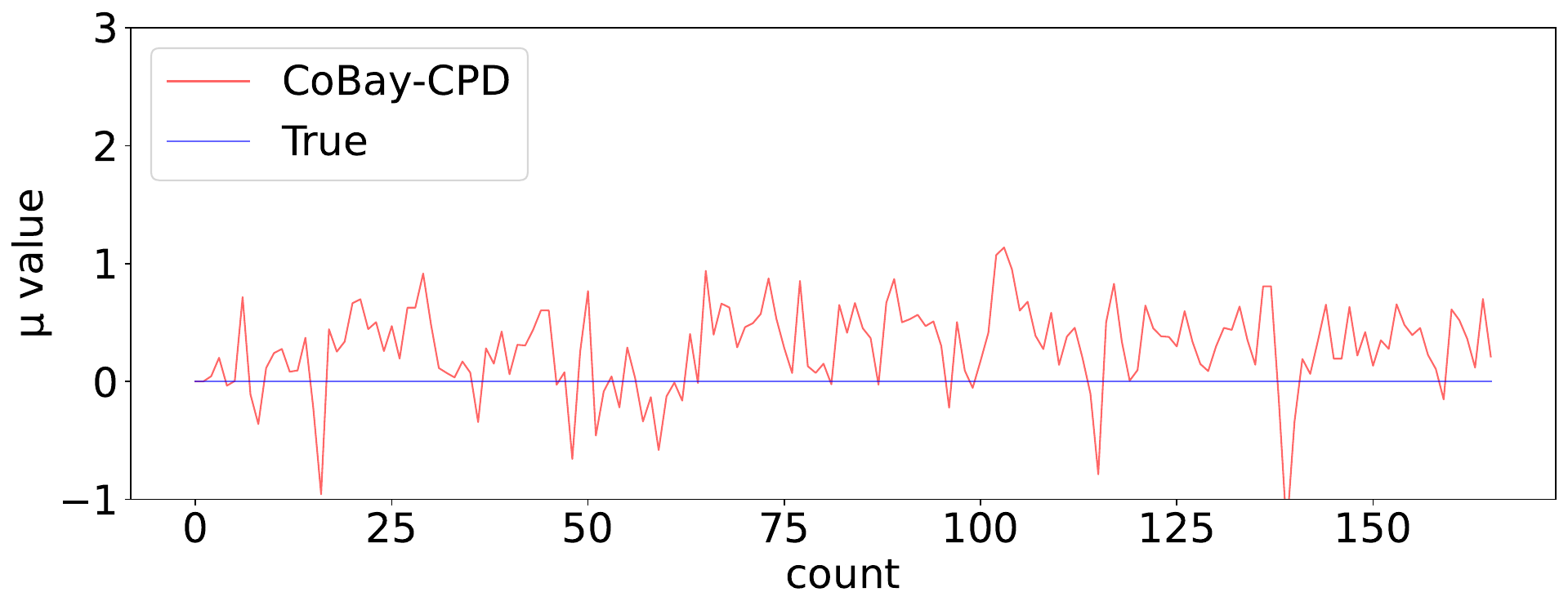}
\subcaption[]{Estimated $\mu$}
\label{fig4a}
\end{minipage}
   \begin{minipage}[b]{0.32\textwidth}
     \centering
     \includegraphics[width=\linewidth]{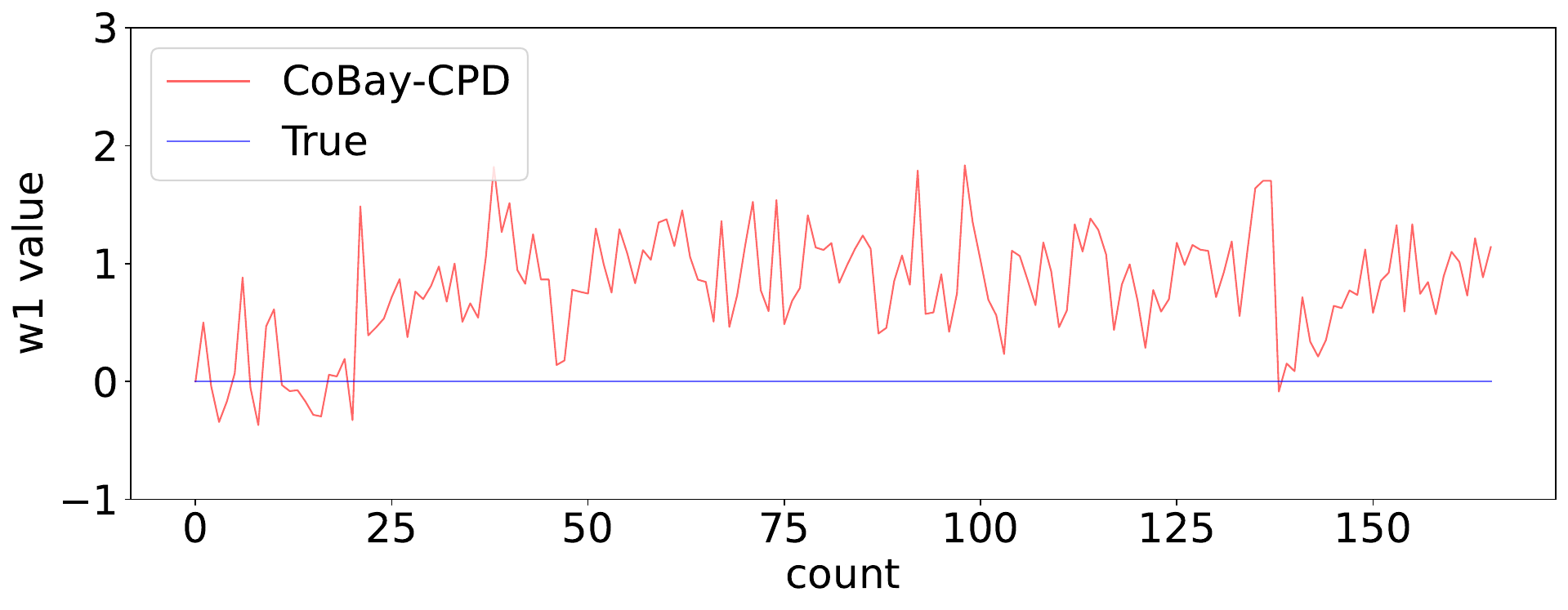}
\subcaption[]{Estimated $w_1$}
\label{fig4b}
\end{minipage}
   \begin{minipage}[b]{0.32\textwidth}
     \centering
     \includegraphics[width=\linewidth]{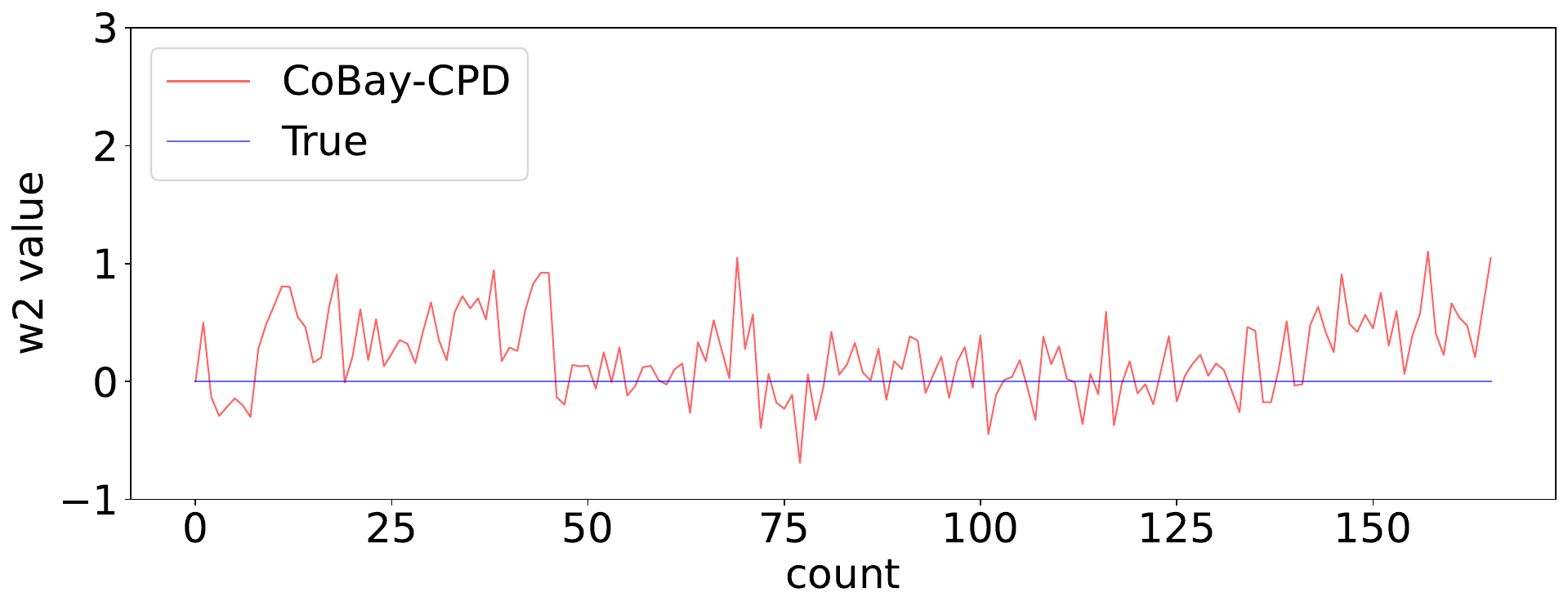}
\subcaption[]{Estimated $w_2$}
\label{fig4c}
\end{minipage}
\begin{minipage}[b]{0.32\textwidth}
\centering
\includegraphics[width=\linewidth]{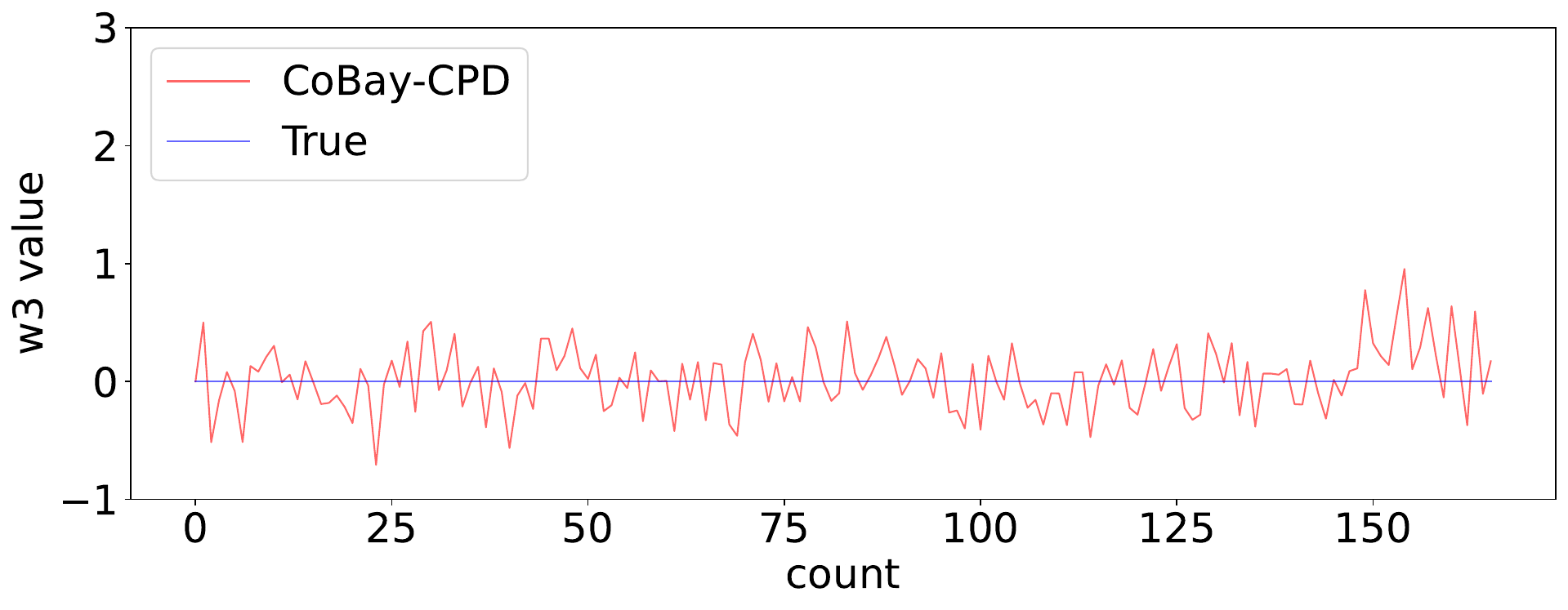}
\subcaption[]{Estimated $w_3$}
\label{fig4d}
\end{minipage}
\begin{minipage}[b]{0.32\textwidth}
\centering
\includegraphics[width=\linewidth]{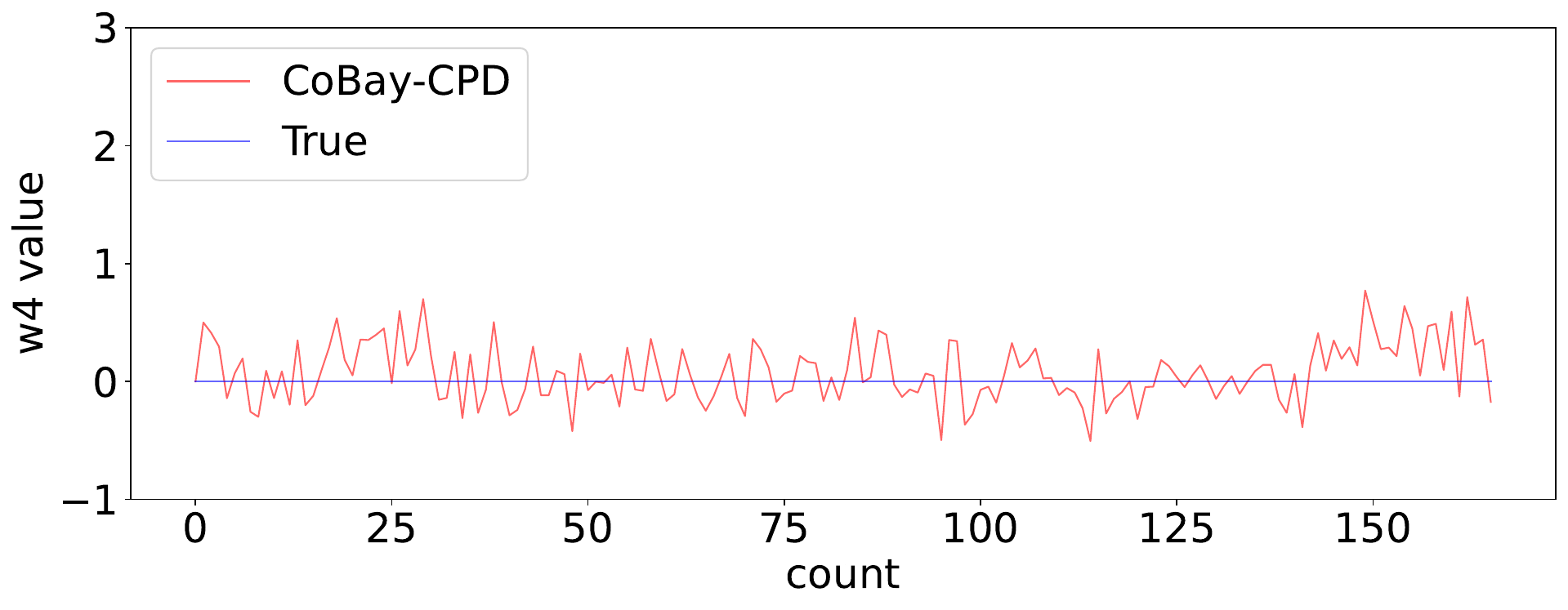}
\subcaption[]{Estimated $w_4$}
\label{fig4e}
\end{minipage}
\caption{Synthetic data: (a) the change point detection results of CoBay-CPD and alternatives, illustrating the change point detection performance; (b) the estimated $\bar{\lambda}$ from CoBay-CPD, indicating the accuracy of parameter estimation of CoBay-CPD;(c)-(g) the estimated parameter (a) $\mu$, (b) $w_1$, (c) $w_2$, (d) $w_3$ and (e) $w_4$ of synthetic data from CoBay-CPD. }
\label{estimated parameter fig}
\end{figure}

In experiment, we assume the basis functions are same as the ground truth, and set the other hyperparameters as 90\% confidence interval and $\mathbf{K}=0.5\mathbf{I}$. The estimation of $\bar{\lambda}$ and the estimated $\mathbf{w}=[\mu,w_{1},\ldots,w_{4}]^\top$ from CoBay-CPD are shown in \cref{estimated parameter fig}. 
We can see that, the estimated parameter  $\bar{\lambda}$ and $\mathbf{w}=[\mu,w_{1},\ldots,w_{4}]^\top$ of synthetic data from CoBay-CPD closely oscillates around the true value, indicating the accuracy of parameter estimation by CoBay-CPD.

\section{Real-world Data Experiment}
\label{app_real_data}
\subsection{Data Processing}
\begin{figure*}[t]
\centering
   \begin{minipage}[b]{0.4\textwidth}
     \centering
     \includegraphics[width=\linewidth]{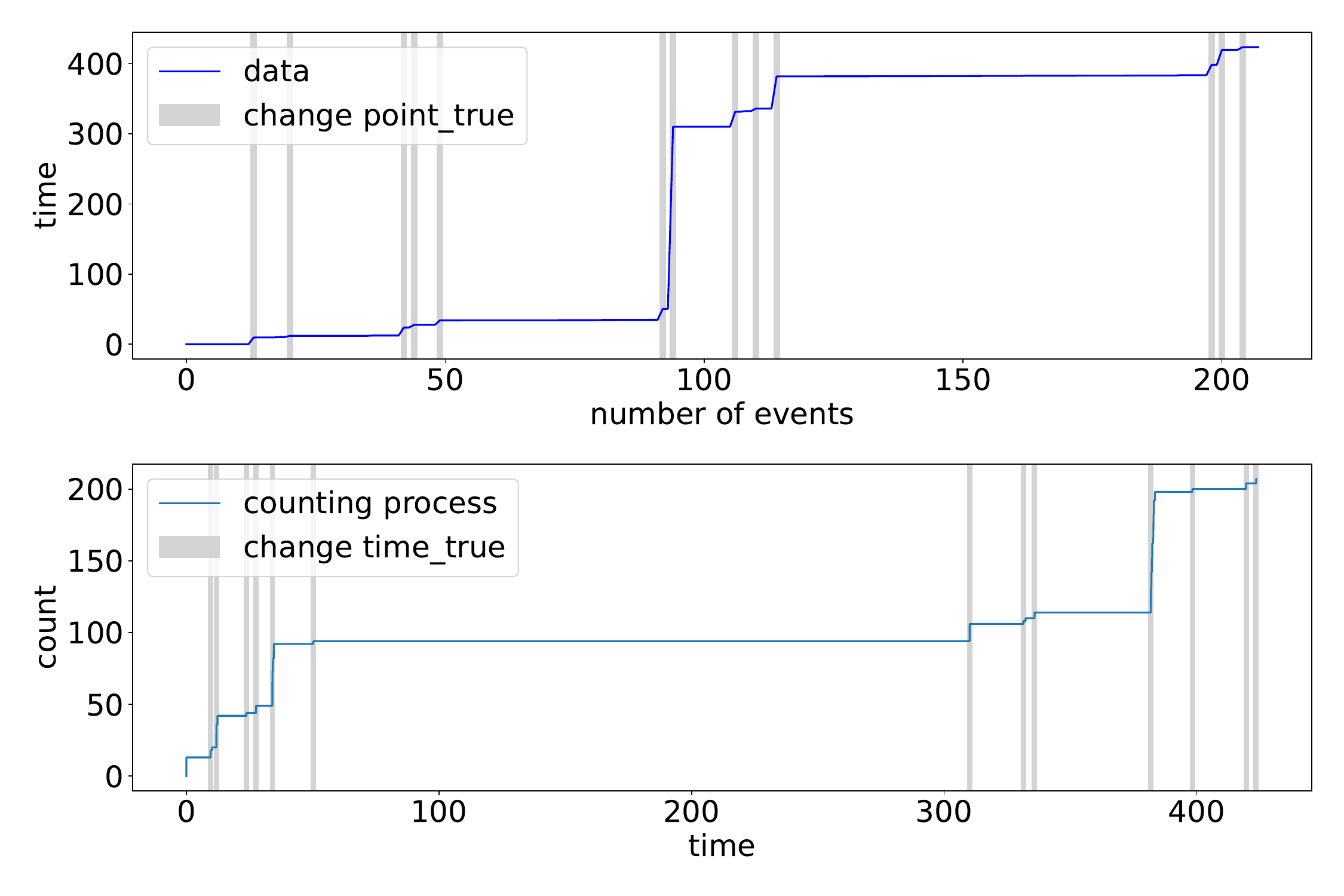}
\subcaption[]{WannaCry Data}
\label{figwan}
\end{minipage}
   \begin{minipage}[b]{0.4\textwidth}
     \centering
     \includegraphics[width=\linewidth]{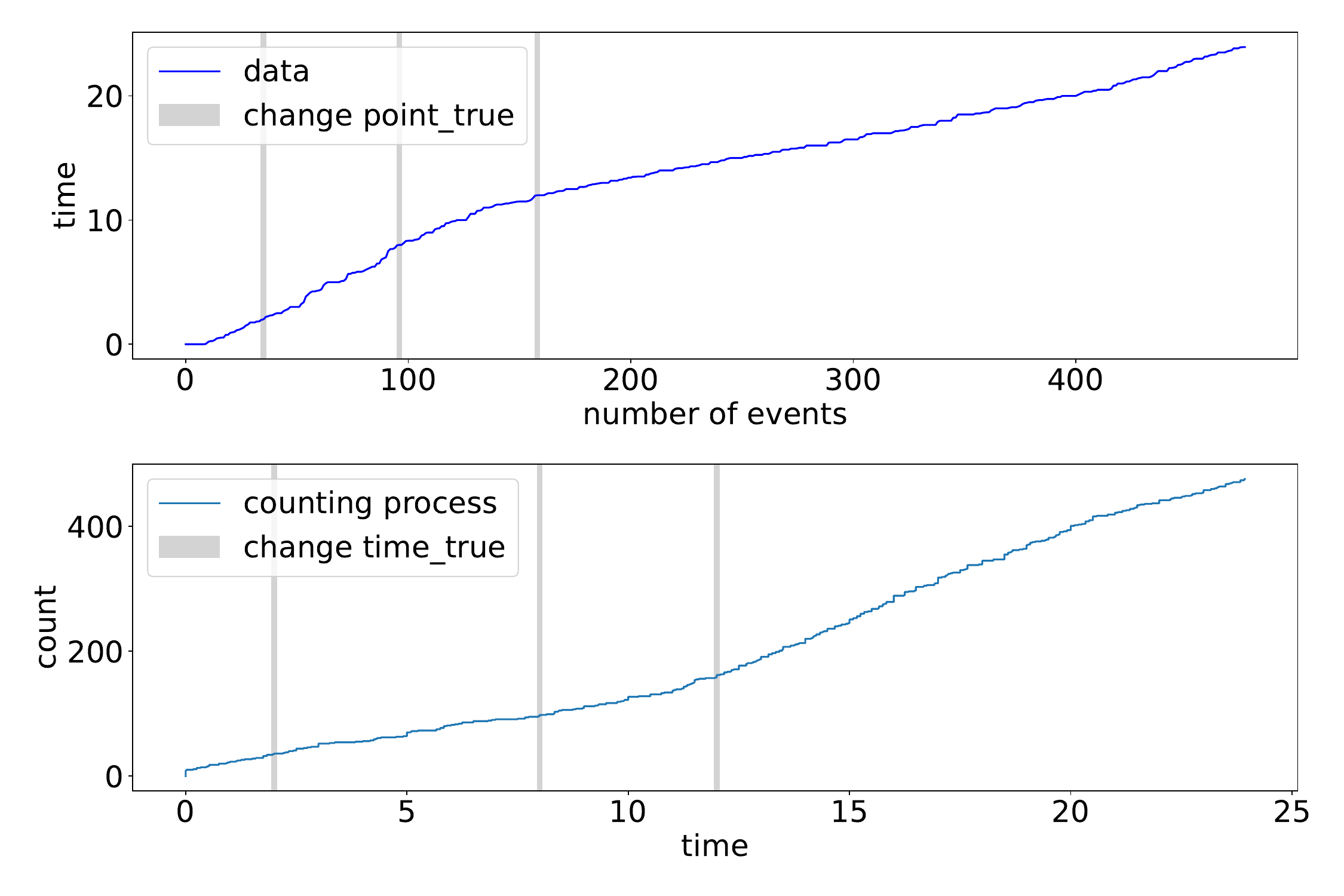}
\subcaption[]{NYC Vehicle Collisions Data}
\label{figcar}
\end{minipage}
\caption{(a) The WannaCry data with ground-truth change points (grey lines). (b) The NYC Vehicle Collisions data with ground-truth change points (grey lines). The upper plot illustrates the increasing of timestamps as events accumulate. The lower plot reverses the axes, representing a counting process.}
\label{fig5pre}
\end{figure*}

The preprocessing details of two real-world datasets are shown below. 
\paragraph{\textbf{WannaCry Cyber Attack}}In May 2017, the WannaCry virus infected more than 200,000 computers worldwide, causing at least hundreds of millions of dollars in damage, and received much attention.  The WannaCry Cyber Attack data contains 208 traffic logs information observations. Each observation contains the relevant timestamp. In this paper,  the points where timestamps surge are taken as the ground truth change points, shown in \cref{figwan}.
\paragraph{\textbf{NYC Vehicle Collisions}}The New York City vehicle collision dataset comprises approximately 1.05 million vehicle collision records, each containing information about the time and location of the collision. For our experiments, we select the records from October 14th, 2017, which contains 477 vehicle collision records. We utilize the change points detected in \cite{zhou2020fast} as the ground truth, shown in \cref{figcar}.

\begin{figure}[t]
    \centering
    \begin{subfigure}[b]{0.40\textwidth}
        \includegraphics[width=\textwidth]{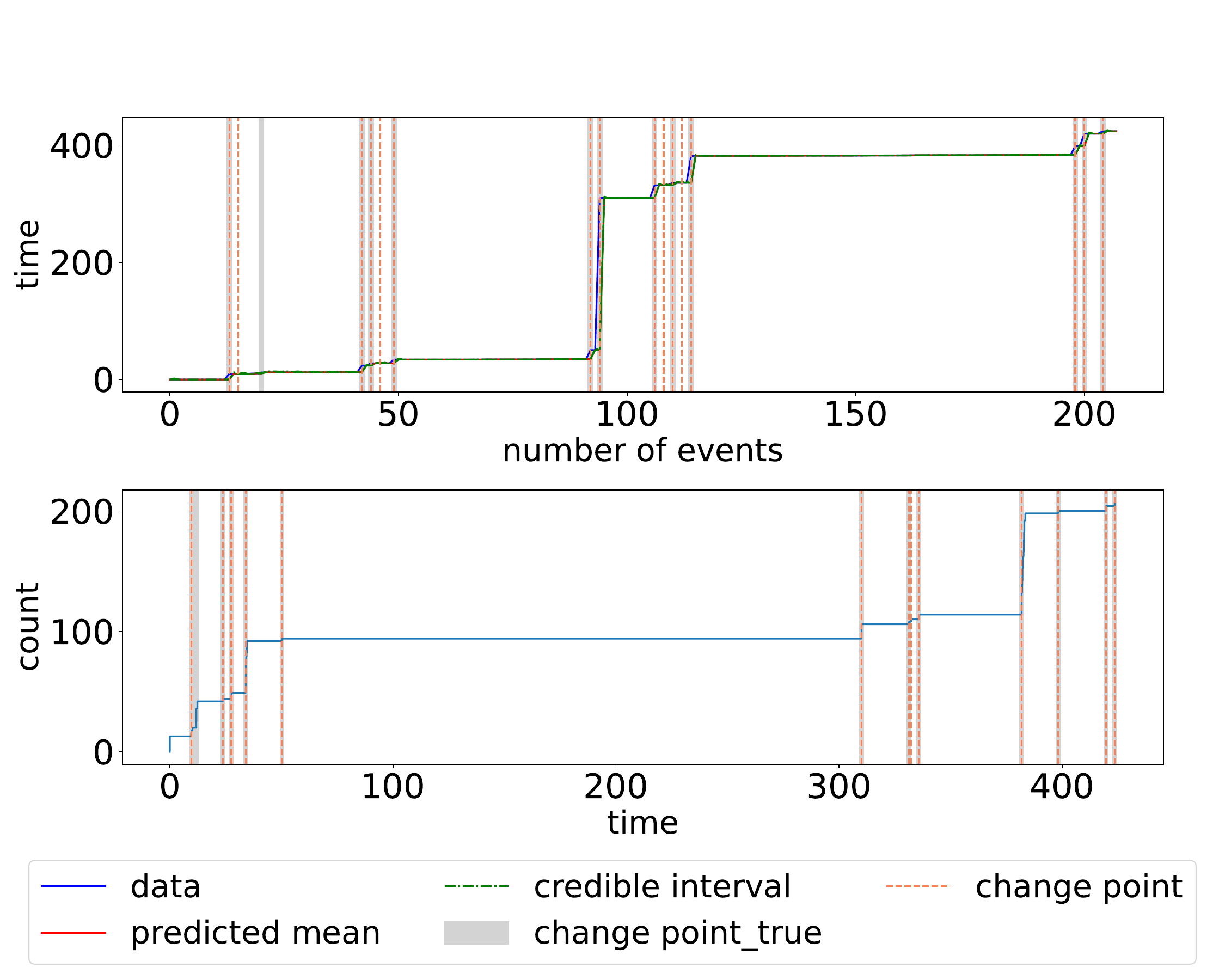}
        \caption{CoBay-CPD}
        \label{fig2b}
    \end{subfigure}
    \begin{subfigure}[b]{0.40\textwidth}
        \includegraphics[width=\textwidth]{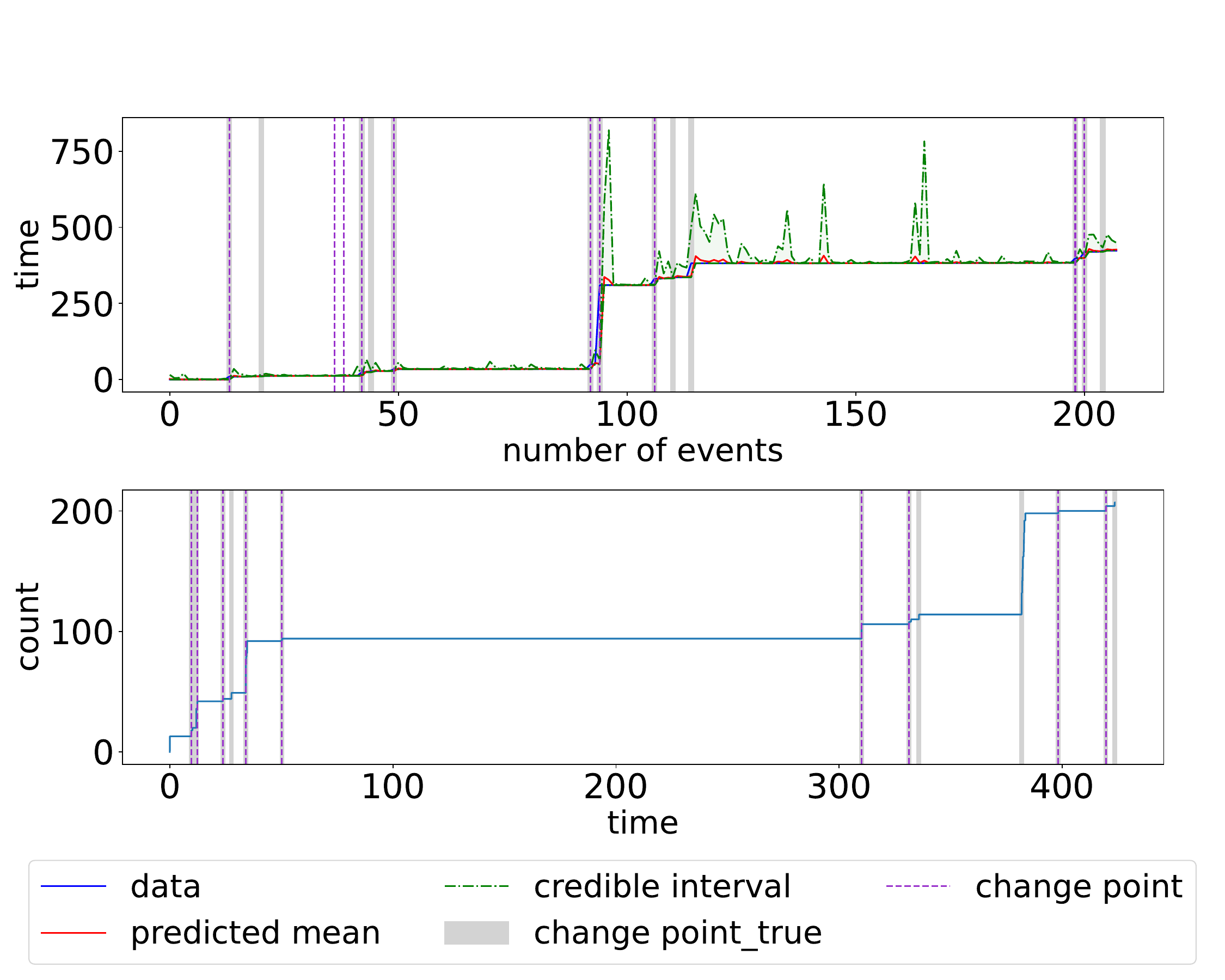}
        \caption{SMCPD}
        \label{fig2c}
    \end{subfigure}
    \centering
    \begin{subfigure}[b]{0.40\textwidth}
        \includegraphics[width=\textwidth]{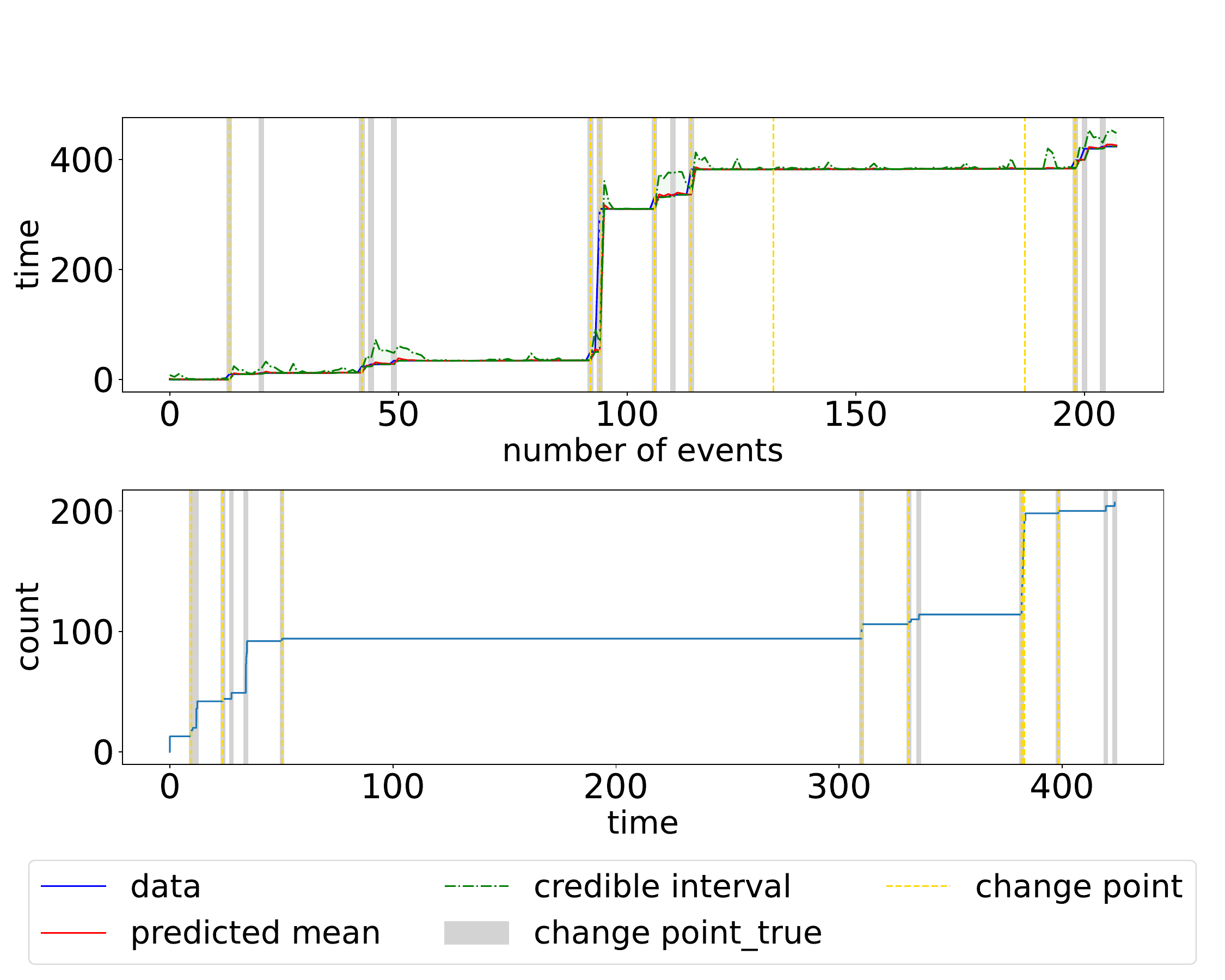}
        \caption{SVCPD}
        \label{fig2d}
    \end{subfigure}
    \begin{subfigure}[b]{0.40\textwidth}
        \includegraphics[width=\textwidth]{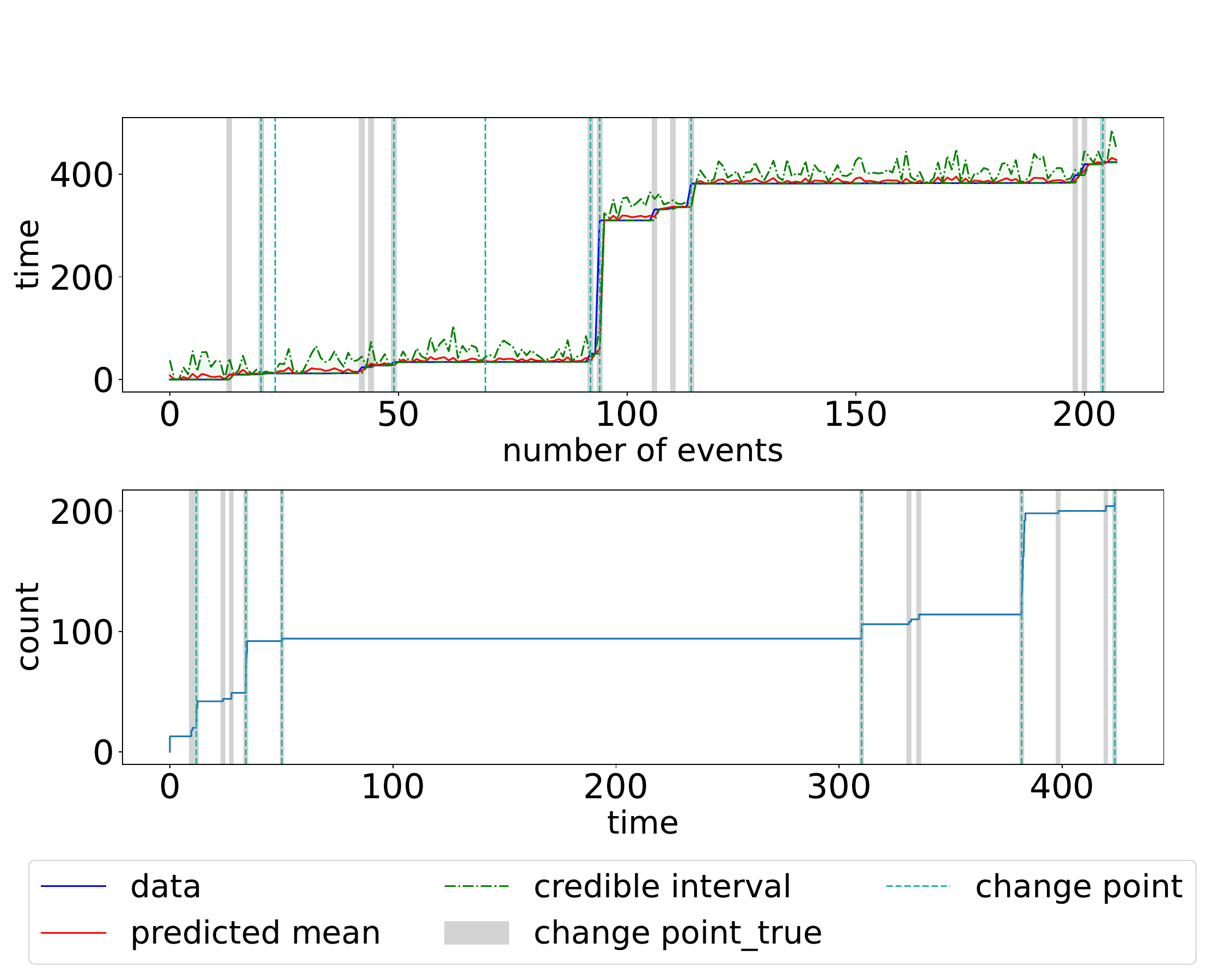}
        \caption{SVCPD+Inhibition}
        \label{fig2e}
    \end{subfigure}
    \caption{The WannaCry data. The upper plot illustrates the increasing of timestamps as events accumulate. The lower plot reverses the axes, representing a counting process. The change point detection result of (a) CoBay-CPD, (b) SMCPD, (c) SVCPD, (d) SVCPD+Inhibiton.}
    \label{fig5}
\end{figure}
\subsection{Results Presentation} \label{app_complete_diagram}
For WannaCry, we adopt a prior distribution $p(\mathbf{w}) = N(\mathbf{w}|\mathbf{0},\mathbf{K})$ for CoBay-CPD, where $\mathbf{K}$ is a diagonal matrix with diagonal entries of $0.5$. Moreover, we choose 90\% confidence interval and $4$ scaled shifted beta densities: $\tilde{\phi}_{1,2,3,4}= \text{Beta}(\tilde{\alpha}=50, \tilde{\beta} = 50, \text{scale} = 6, \text{shift} = \{-2,-1,0,1\})$ as basis functions. 
The complete graph of  experimental results of four methods on WannaCry Cyber Attack Dataset is shown in \cref{fig5}.
\Cref{fig2b,fig2c,fig2d,fig2e} display the change point detection outcomes of different methods applied to WannaCry data.
The blue line is the real data, the red solid line is the mean of the predicted points, the green dotted line is the confidence interval, and the orange, purple, yellow and turquoise line are the detected change point location.

\begin{figure*}[t]
\centering
   \begin{minipage}[b]{0.4\textwidth}
     \centering
     \includegraphics[width=\linewidth]{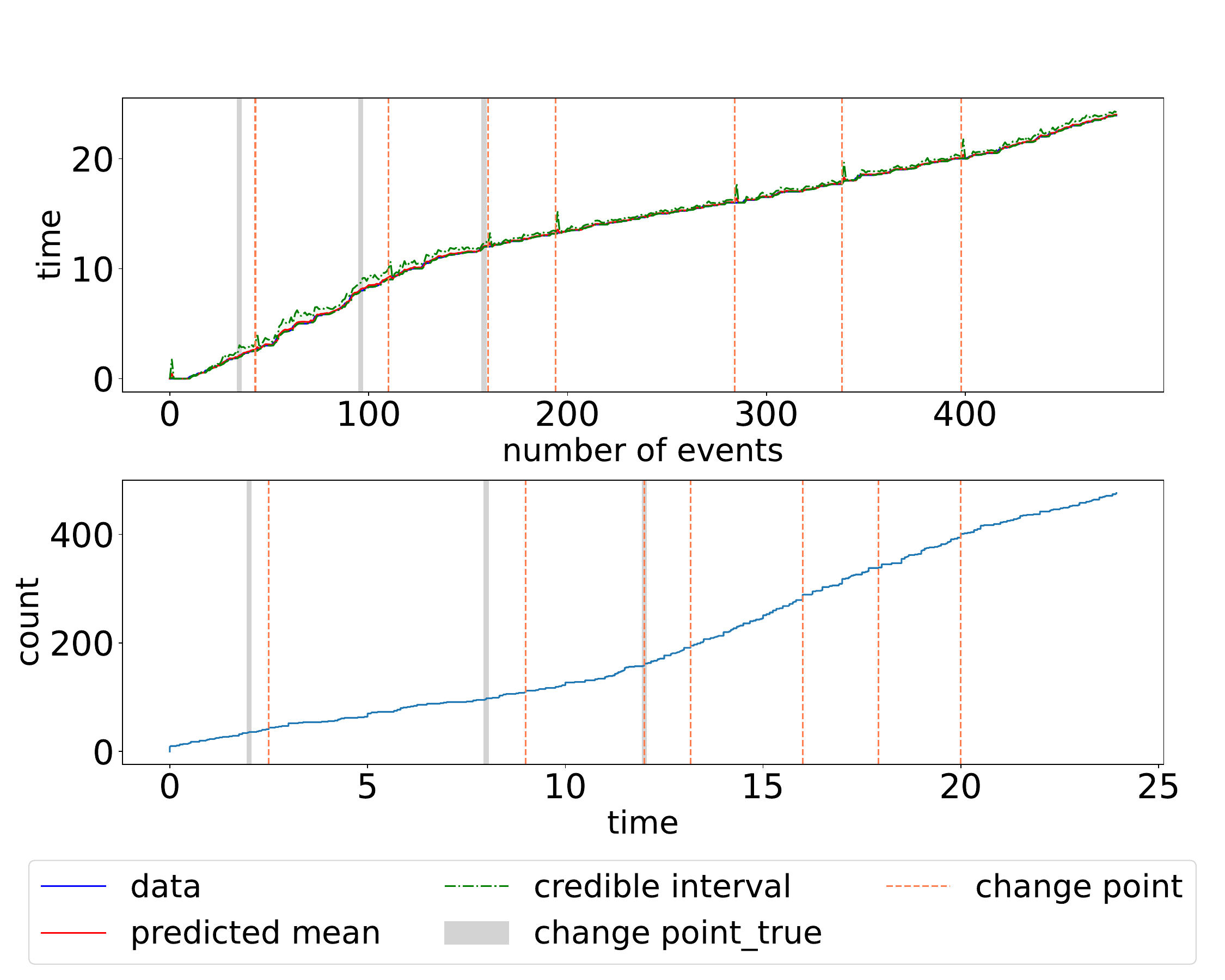}
\subcaption[]{CoBay-CPD}
\label{fig3b}
\end{minipage}
   \begin{minipage}[b]{0.4\textwidth}
     \centering
     \includegraphics[width=\linewidth]{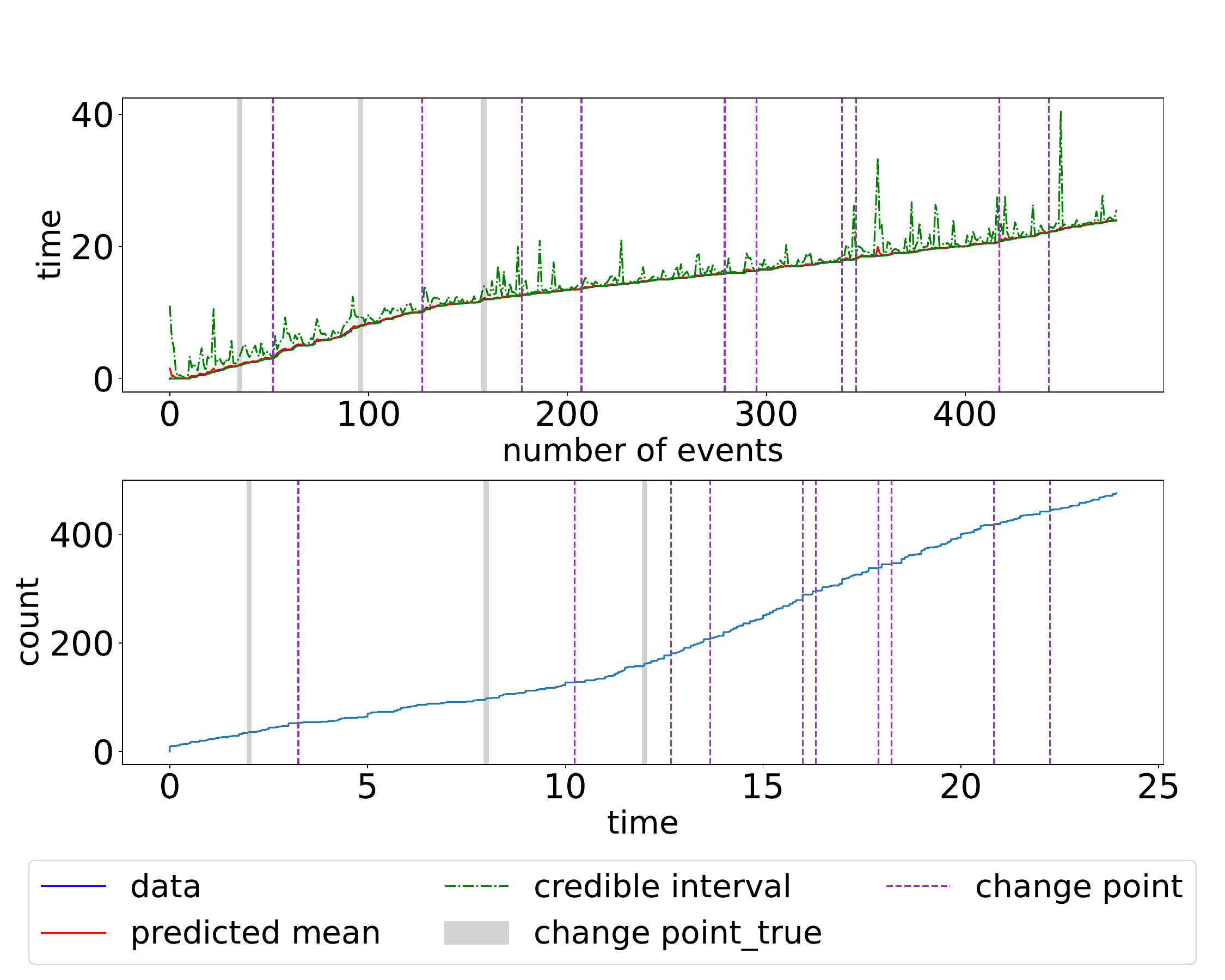}
\subcaption[]{SMCPD}
\label{fig3c}
\end{minipage}
   \begin{minipage}[b]{0.4\textwidth}
     \centering
     \includegraphics[width=\linewidth]{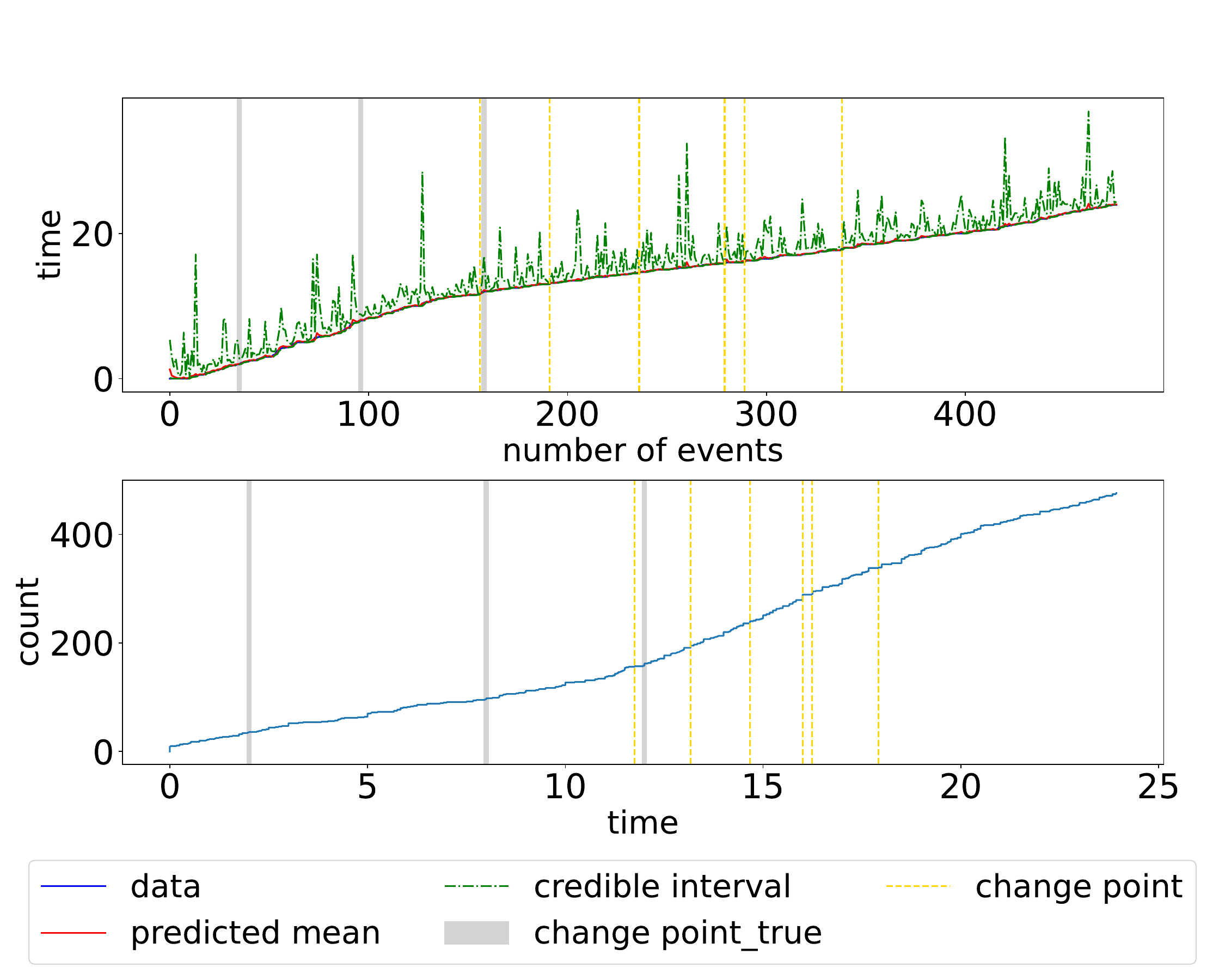}
\subcaption[]{SVCPD}
\label{fig3d}
\end{minipage}
\begin{minipage}[b]{0.4\textwidth}
\centering
\includegraphics[width=\linewidth]{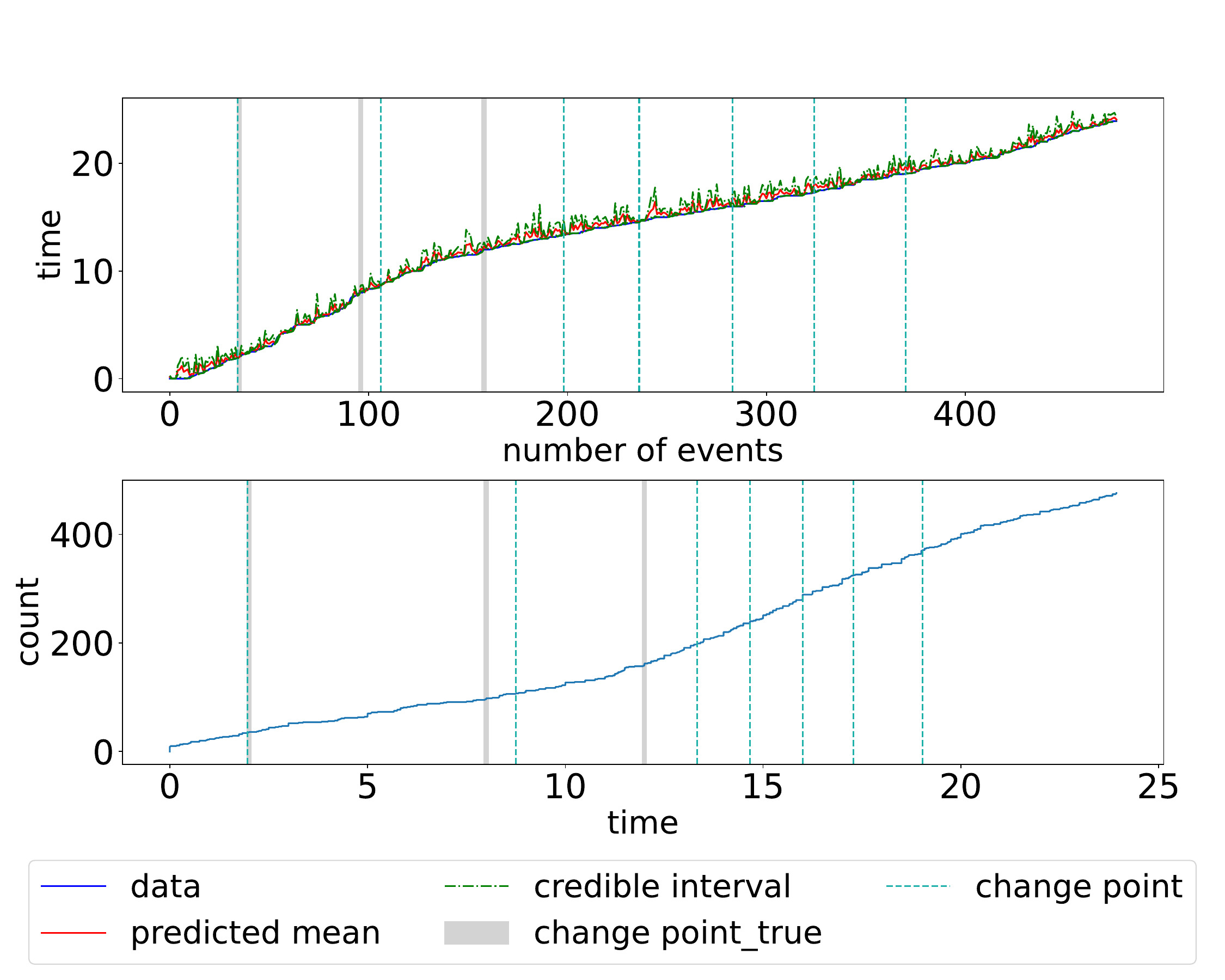}
\subcaption[]{SVCPD+Inhibiton}
\label{fig3e}
\end{minipage}
\caption{The NYC Vehicle Collisions data. The upper plot illustrates the increasing of timestamps as events accumulate. The lower plot reverses the axes, representing a counting process. The change point detection result of (a) CoBay-CPD, (b) SMCPD, (c) SVCPD, (d) SVCPD+Inhibiton.}
\label{fig6}
\end{figure*}

For NYC Vehicle Collisions, we choose $4$ scaled shifted beta densities: $\tilde{\phi}_{1,2,3,4}= \text{Beta}(\tilde{\alpha}=10, \tilde{\beta} = 30, \text{scale} = 6, \text{shift} = \{-2,-1,0,1\})$ as basis functions, 90\% confidence interval and $\mathbf{K}=0.5\mathbf{I}$, which is same as that in WannaCry. 
The change points in the NYC Vehicle Collisions are not as obvious as those in the WannaCry, making change point detection a more challenging task for this dataset. 
The complete graph of experimental results of four methods on NYC Vehicle Collisions Dataset is shown in \cref{fig6}.
The blue line is the real data, the red solid line is the mean of the predicted points, the green dotted line is the confidence interval, and the orange, purple, yellow and turquoise line are the detected change point location. 
\Cref{fig3b,fig3c,fig3d,fig3e} show the change point detection outcomes of four methods for the NYC data. 
Notably, SVCPD detects fewer change points, while SMCPD identify an excessive number.

\section{Stress Tests}
\label{ST}
\subsection{Test 1: Number of Change Points}
We conduct a stress test with the number of change points: $1,2,3$. 
We generate three sets of synthetic data concatenated by some segments of Hawkes process data. 
In all these segments of Hawkes process data, we assume $4$ scaled beta densities: $\tilde{\phi}_{1,2,3,4}= \text{Beta}(\tilde{\alpha}=50, \tilde{\beta} = 50, \text{scale} = 6, \text{shift} = \{-2,-1,0,1\})$ as the basis functions with support $[0, T_{\phi}=6]$ and $\mu = 0$ as the baseline activation. 
However, they have different intensity upperbounds.
For $\#\ \text{of change points} = 1$, we let $\bar{\lambda}_{11} = 5$, $\bar{\lambda}_{12} = 10$; For $\#\ \text{of change points} = 2$, we let $\bar{\lambda}_{21} = 5$, $\bar{\lambda}_{22} = 10$, and $\bar{\lambda}_{23} = 3$; For $\#\ \text{of change points} = 3$, we let $\bar{\lambda}_{31} = 5$, $\bar{\lambda}_{32} = 10$, $\bar{\lambda}_{33} = 3$, and $\bar{\lambda}_{34} = 8$. We use the thinning algorithm to generate these sequences according to the intensity specified above.

\subsection{Test 2: Difference between Adjacent Parameters}
We conduct a stress test with $\Delta \bar{\lambda} = 0.1, 1, 5$. We generate three sets of synthetic data by concatenating two segments of Hawkes process data. Within each segment, we assume four scaled beta densities: $\tilde{\phi}_{1,2,3,4}= \text{Beta}(\tilde{\alpha}=50, \tilde{\beta}=50, \text{scale}=6, \text{shift}={-2,-1,0,1})$ as the basis functions with support $[0, T_{\phi}=6]$, and $\mu = 0$ as the baseline activation. However, they possess different intensity upper bounds. Specifically, for $\Delta \bar{\lambda} = 0.1$, we set $\bar{\lambda}_{11} = 10$ and $\bar{\lambda}_{12} = 10.1$; for $\Delta \bar{\lambda} = 1$, we set $\bar{\lambda}_{21} = 10$ and $\bar{\lambda}_{22} = 9$; for $\Delta \bar{\lambda} = 5$, we set $\bar{\lambda}_{31} = 10$ and $\bar{\lambda}_{32} = 5$. We use the thinning algorithm to generate these sequences according to the intensity specified above. 

\subsection{Test 3: Closeness between Adjacent Change Points}
We conduct a stress test with $\Delta_t = 5, 10, 15$. Three sets of synthetic data are generated by concatenating three segments of Hawkes process data. Within each segment, we assume four scaled beta densities: $\tilde{\phi}_{1,2,3,4} = \text{Beta}(\tilde{\alpha}=50, \tilde{\beta}=50, \text{scale}=6, \text{shift}={-2,-1,0,1})$ as the basis functions with support $[0, T_{\phi}=6]$, and $\mu = 0$ as the baseline activation. 
They have different intensity upperbounds $\bar{\lambda}_{1} = 10$, $\bar{\lambda}_{2} = 5$, and $\bar{\lambda}_{3} = 15$. 
We use the thinning algorithm to generate these sequences according to the intensity specified above. 
We adjust the data length of the second segment from 5 to 10 to 15, thereby controlling the interval $\Delta_t$ between two adjacent change points from 5 to 10 to 15.

\subsection{Stree Tests Supplements}
Due to the page limit, we only presented the stress test results for our own method in the main paper. However, based on a suggestion from an anonymous reviewer to include the stress test results for the baselines, we have provided them in \cref{table5,table6,table7}.
\begin{table}[!h]
\caption{The FNR, FPR and MSE of CoBay-CPD and other baselines on synthetic dataset with different number of change points.}
\label{table5}
\scalebox{0.67}{
\begin{tabular}{@{}c|ccc|ccc|ccc@{}}
\toprule
\multirow{2}{*}{Model} & \multicolumn{3}{c|}{1}                                                         & \multicolumn{3}{c|}{2}                                                         & \multicolumn{3}{c}{3}                                                          \\ \cmidrule(l){2-10} 
                       & FNR($\downarrow$)        & FPR(\% $\downarrow$)     & MSE($\downarrow$)        & FNR($\downarrow$)        & FPR(\% $\downarrow$)     & MSE($\downarrow$)        & FNR($\downarrow$)        & FPR(\% $\downarrow$)     & MSE($\downarrow$)        \\ \midrule
SMCPD                  & 0.33 $\pm$ 0.47          & 0.63 $\pm$ 0.63          & 0.08 $\pm$ 0.03          & 0.38 $\pm$ 0.41          & 0.76 $\pm$ 0.26          & 0.07 $\pm$ 0.01          & 0.67 $\pm$ 0.19          & 1.23 $\pm$ 0.35          & 0.08 $\pm$ 0.01          \\
SVCPD                  & 0.67 $\pm$ 0.47          & 0.63 $\pm$ 0.95          & 0.06 $\pm$ 0.01          & 0.50 $\pm$ 0.35          & 0.76 $\pm$ 0.26          & 0.06 $\pm$ 0.00          & 0.50 $\pm$ 0.32          & 1.74 $\pm$ 0.65          & 0.15 $\pm$ 0.05          \\
SVCPD+Inhi             & 0.33 $\pm$ 0.47          & 1.88 $\pm$ 0.63          & 0.08 $\pm$ 0.01          & 0.33 $\pm$ 0.24          & 0.60 $\pm$ 0.00          & 0.16 $\pm$ 0.01          & 0.28 $\pm$ 0.23          & 1.84 $\pm$ 0.50          & 0.09 $\pm$ 0.00          \\
CoBay-CPD              & \textbf{0.00 $\pm$ 0.00} & \textbf{0.43 $\pm$ 0.60} & \textbf{0.04 $\pm$ 0.00} & \textbf{0.13 $\pm$ 0.22} & \textbf{0.46 $\pm$ 0.26} & \textbf{0.05 $\pm$ 0.00} & \textbf{0.11 $\pm$ 0.14} & \textbf{0.31 $\pm$ 0.50} & \textbf{0.07 $\pm$ 0.01} \\ \bottomrule
\end{tabular}
}
\end{table}

\begin{table}[!h]
\caption{The FNR, FPR and MSE of CoBay-CPD and other baselines on synthetic dataset with different difference between adjacent $\bar\lambda$'s ($\Delta \bar{\lambda}$).}
\label{table6}
\scalebox{0.67}{
\begin{tabular}{@{}c|ccc|ccc|ccc@{}}
\toprule
\multirow{2}{*}{Model} & \multicolumn{3}{c|}{0.1}                                                       & \multicolumn{3}{c|}{1}                                                         & \multicolumn{3}{c}{5}                                                          \\ \cmidrule(l){2-10} 
                       & FNR($\downarrow$)        & FPR(\% $\downarrow$)     & MSE($\downarrow$)        & FNR($\downarrow$)        & FPR(\% $\downarrow$)     & MSE($\downarrow$)        & FNR($\downarrow$)        & FPR(\% $\downarrow$)     & MSE($\downarrow$)        \\ \midrule
SMCPD                  & 1.00 $\pm$ 0.00          & \textbf{1.20 $\pm$ 0.00} & 0.06 $\pm$ 0.01          & 0.50 $\pm$ 0.50          & 0.70 $\pm$ 0.70          & 0.06 $\pm$ 0.02          & 0.33 $\pm$ 0.47          & 0.63 $\pm$ 0.63          & 0.08 $\pm$ 0.03          \\
SVCPD                  & 1.00 $\pm$ 0.00          & 2.41 $\pm$ 0.98          & 0.05 $\pm$ 0.01          & 0.83 $\pm$ 0.37          & 3.29 $\pm$ 1.05          & 0.06 $\pm$ 0.01          & 0.67 $\pm$ 0.47          & 0.63 $\pm$ 0.95          & 0.06 $\pm$ 0.01          \\
SVCPD+Inhi             & \textbf{0.67 $\pm$ 0.47} & 1.41 $\pm$ 0.83          & 0.06 $\pm$ 0.00          & 0.33 $\pm$ 0.47          & 1.17 $\pm$ 0.52          & 0.06 $\pm$ 0.00          & 0.33 $\pm$ 0.47          & 1.88 $\pm$ 0.63          & 0.08 $\pm$ 0.01          \\
CoBay-CPD              & 1.00 $\pm$ 0.00          & 1.61 $\pm$ 0.57          & \textbf{0.02 $\pm$ 0.00} & \textbf{0.25 $\pm$ 0.43} & \textbf{0.35 $\pm$ 0.60} & \textbf{0.03 $\pm$ 0.00} & \textbf{0.00 $\pm$ 0.00} & \textbf{0.43 $\pm$ 0.60} & \textbf{0.04 $\pm$ 0.00} \\ \bottomrule
\end{tabular}
}
\end{table}

\begin{table}[!h]
\caption{The FNR, FPR and MSE of CoBay-CPD and other baselines on synthetic dataset with different closeness between two change points ($\Delta t$).}
\label{table7}
\scalebox{0.67}{
\begin{tabular}{@{}c|ccc|ccc|ccc@{}}
\toprule
\multirow{2}{*}{Model} & \multicolumn{3}{c|}{5}                                                         & \multicolumn{3}{c|}{10}                                                        & \multicolumn{3}{c}{15}                                                                       \\ \cmidrule(l){2-10} 
                       & FNR($\downarrow$)        & FPR(\% $\downarrow$)     & MSE($\downarrow$)        & FNR($\downarrow$)        & FPR(\% $\downarrow$)     & MSE($\downarrow$)        & FNR($\downarrow$)        & FPR(\% $\downarrow$)     & \multicolumn{1}{c}{MSE($\downarrow$)} \\ \midrule
SMCPD                  & 0.42 $\pm$ 0.34          & \textbf{0.75 $\pm$ 0.75} & 0.03 $\pm$ 0.01          & 0.67 $\pm$ 0.24          & \textbf{0.23 $\pm$ 0.52}          & 0.05 $\pm$ 0.01          & 0.17 $\pm$ 0.24          & 1.00 $\pm$ 0.83          & 0.07 $\pm$ 0.01                        \\
SVCPD                  & 0.42 $\pm$ 0.19          & 1.24 $\pm$ 0.56          & 0.03 $\pm$ 0.01          & 0.75 $\pm$ 0.25          & 0.46 $\pm$ 0.65          & 0.05 $\pm$ 0.01          & \textbf{0.08 $\pm$ 0.19} & 3.01 $\pm$ 1.15          & 0.06 $\pm$ 0.01                        \\
SVCPD+Inhi             & 0.58 $\pm$ 0.19          & 1.24 $\pm$ 1.33          & 0.05 $\pm$ 0.01          & 0.25 $\pm$ 0.38          & \textbf{0.23 $\pm$ 0.52}          & 0.05 $\pm$ 0.00          & 0.17 $\pm$ 0.24          & 2.01 $\pm$ 1.33          & 0.06 $\pm$ 0.00                        \\
CoBay-CPD              & \textbf{0.33 $\pm$ 0.24} & 1.00 $\pm$ 0.70          & \textbf{0.01 $\pm$ 0.00} & \textbf{0.00 $\pm$ 0.00} & 0.93 $\pm$ 0.65 & \textbf{0.02 $\pm$ 0.00} & \textbf{0.08 $\pm$ 0.19} & \textbf{0.80 $\pm$ 0.57} & \textbf{0.03 $\pm$ 0.00}               \\ \bottomrule
\end{tabular}
}
\end{table}

\end{document}